\documentclass{article} % For LaTeX2e
\usepackage{iclr2027_conference,times}

\usepackage{amsmath,amsfonts,bm}

\def\eqref#1{equation~\ref{#1}}
\def\1{\bm{1}}

\DeclareMathAlphabet{\mathsfit}{\encodingdefault}{\sfdefault}{m}{sl}
\SetMathAlphabet{\mathsfit}{bold}{\encodingdefault}{\sfdefault}{bx}{n}

\newcommand{\eat}[1]{}

\usepackage{balance}
\usepackage{times}
\usepackage{latexsym}
\usepackage{amsfonts}
\usepackage{amsmath}
\usepackage{colortbl}
\usepackage{epsfig}
\usepackage{xspace}
\usepackage{xcolor}
\usepackage{graphicx}
\usepackage{comment}
\usepackage{booktabs}
\usepackage{multirow}
\usepackage{framed}
\usepackage{subcaption}
\usepackage{wrapfig}
\usepackage{circledsteps}
\usepackage[utf8]{inputenc}
\usepackage{microtype}
\usepackage{inconsolata}
\usepackage[colorlinks=false, linkcolor=black, urlcolor=magenta, citecolor=black]{hyperref}
\usepackage{siunitx}
\usepackage{titlesec}
\usepackage{placeins}

\usepackage[ruled,linesnumbered]{algorithm2e}
\usepackage{algpseudocode}

\algnewcommand\algorithmicinput{\textbf{Input:}}
\algnewcommand\algorithmicoutput{\textbf{Output:}}
\algnewcommand\Input{\item[\algorithmicinput]}
\algnewcommand\Output{\item[\algorithmicoutput]}

\algdef{SE}[DOWHILE]{Do}{doWhile}{\algorithmicdo}[1]{\algorithmicwhile\ #1}%
\usepackage{lineno}
\definecolor{green}{RGB}{0,128,0}

\definecolor{yellow}{RGB}{255,200,18}

\newcommand{\bi}{\begin{itemize}}
\newcommand{\ei}{\end{itemize}}

\newcommand{\be}{\begin{enumerate}}
\newcommand{\ee}{\end{enumerate}}
\newcommand{\beqn}{\begin{eqnarray*}}
\newcommand{\eeqn}{\end{eqnarray*}}

\newcommand{\eg}{{\em e.g.,}\xspace}

\newcommand{\actiontag}[2]{%
    \begingroup
    \setlength{\fboxsep}{1.5pt}%
    \fcolorbox{#1}{white}{%
        \textcolor{#1}{\texttt{<#2>}}%
    }%
    \endgroup
}
\newcommand{\framework}[1]{{\scriptsize #1}}

\usepackage{tabu}                      % only used for the table example
\usepackage{booktabs}                  % only used for the table example
\usepackage{lipsum}                    % used to generate placeholder text
\usepackage{mwe}                       % used to generate placeholder figures

\usepackage{listings}
\usepackage{textcomp} % provides \textquotedbl for listings in OT1/Times
\usepackage{pifont}
\usepackage{tikz}
\usepackage{enumitem}
\usepackage{natbib}
\usepackage{utfsym} % For checkmarks and cross symbols
\usepackage[most]{tcolorbox}

\usetikzlibrary{shapes,snakes}
\usetikzlibrary{calc}

\definecolor{shadecolor}{RGB}{220,220,220}

\tikzstyle{mybox} = [draw=black, fill=black!5, thick,
    rectangle, rounded corners, inner sep=2pt, inner ysep=8pt]
\tikzstyle{fancytitle} =[fill=black, text=white]

\tcbset{
  stagebox/.style 2 args={% #1=背景色, #2=文字色
    enhanced,
    on line,
    valign=center,
    colback=#1,
    coltext=#2,
    arc=2pt,
    boxrule=0pt,
    boxsep=0pt,
    left=1pt,
    right=1pt,
    top=1.25pt,
    bottom=1pt,
    fontupper=\sffamily\bfseries\footnotesize,
  }
}

\newcommand{\sys}{\textsc{DocPilot}\xspace}

\definecolor{mygreen}{RGB}{203,230,204}
\definecolor{mydarkgreen}{RGB}{100,230,100}
\definecolor{myred}{RGB}{255,182,193}
\definecolor{mygray}{RGB}{230,230,230}
\definecolor{myblue}{RGB}{138,180,189}
\definecolor{myyellow}{RGB}{255,185,84}
\definecolor{radargreen}{HTML}{38ADA9}

\definecolor{stageAcolor}{HTML}{90b3bd}
\definecolor{stageBcolor}{HTML}{738348}

\usepackage{xcolor}

\definecolor{DeltaUpBg}{HTML}{E6F4EA}
\definecolor{DeltaUpFg}{HTML}{137333}
\definecolor{DeltaDnBg}{HTML}{FDE7E9}
\definecolor{DeltaDnFg}{HTML}{A50E0E}

\definecolor{firstBest}{HTML}{B2D8F2}
\definecolor{secondBest}{HTML}{FCE5E3}

\definecolor{navigatecolor}{HTML}{5B8DB8} % muted blue
\definecolor{extractcolor}{HTML}{6FA89A}  % muted teal
\definecolor{focuscolor}{HTML}{8B79B5}    % muted purple
\definecolor{solvecolor}{HTML}{C39B57}    % muted gold
\definecolor{analyzecolor}{HTML}{C87575}  % muted red
\definecolor{answercolor}{HTML}{70859A}   % blue gray

\newcommand{\cmark}{\textcolor{green!60!black}{\ding{51}}}
\newcommand{\xmark}{\textcolor{red}{\ding{55}}}

\usepackage{url}

\title{DocPILOT: Interactive Agent Orchestration and Training for Long Visual Document Understanding}
\author{
Zhuowen Liang$^{1}$,
Xiaotian Lin$^{1}$,
Yuyu Luo$^{1}$,
Haixun Wang$^{2}$,
Nan Tang$^{1} \thanks{Corresponding author: Nan Tang (E-mail: nantang@hkust-gz.edu.cn)}$
 \vspace{.5em} 
  \\
$^{1}$The Hong Kong University of Science and Technology (Guangzhou),  $^{2}$EvenUp, USA \\
}

\iclrfinalcopy % Uncomment for camera-ready version, but NOT for submission.
\begin{document}

\maketitle
\vspace{-1em}

\begin{figure}[h!]
   \vspace{-1em}
   \centering
   \includegraphics[width=\linewidth]{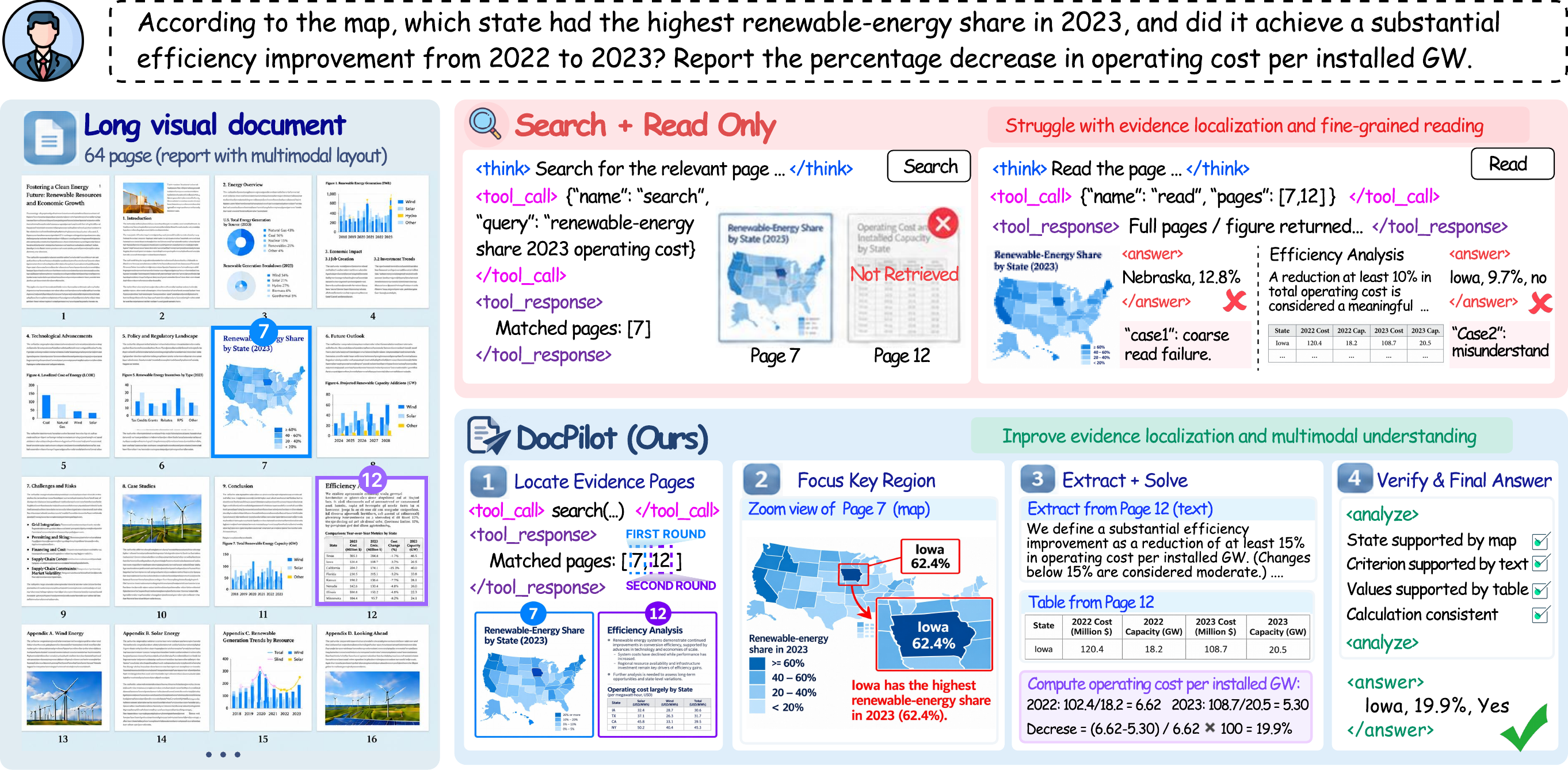}
   \vspace{-2em}
   \caption{
       Illustration of \sys's capabilities in evidence seeking and multimodal understanding.
   }
   \label{fig:teaser}
\end{figure}

\begin{abstract}
   Multimodal long-document question answering requires models to seek sparse evidence across pages and modalities, understand heterogeneous document content, and reason over the collected evidence.
   Even advanced vision-language models (VLMs) can overlook sparse evidence and struggle with cross-modal reasoning in a single pass.
   Existing document agents further introduce multi-step interaction, but some still rely on predefined workflows or coarse-grained actions, limiting the flexibility and granularity of document reasoning.
   We introduce \textbf{\sys}, an \emph{agentic document framework} that supports expressive orchestration and makes complex document solving learnable by small language models (SLMs).
    \sys orchestrates evidence seeking through page-level navigation and fine-grained evidence localization, and integrates local visual inspection, code-based numerical reasoning, and cross-modal reasoning for multimodal understanding.
   To transfer these capabilities to small models, we propose \emph{curriculum-based agentic training} that progresses from single-ability learning to adaptive orchestration through supervision from data-grounded reasoning traces and executable interaction trajectories.
   Across MMLongBench-Doc, DocBench, FinRAGBench-V, and MADQA, \sys achieves strong gains with frontier VLM backbones, while \sys-4B delivers competitive performance against substantially larger models, demonstrating strong effectiveness and generalization across diverse long-document reasoning settings.
   The code is available at \url{https://github.com/HKUSTDial/DocPilot}.
\end{abstract}
\vspace{-1em}

\section{Introduction}
\label{sec:introduction}

Multimodal long-document question answering (DocVQA) is a fundamental problem in real-world document intelligence, with applications in financial analysis, scientific literature, legal or compliance review, and enterprise knowledge management~\citep{liu2025comprehensive, DBLP:conf/wacv/MathewKJ21}.
Unlike text-only QA or single-image understanding, real documents contain heterogeneous information sources, including paragraphs, tables, charts, figures, and complex layouts.
Answering questions over such documents often requires connecting evidence distributed across multiple pages/documents and modalities, remaining challenging even for advanced vision-language models (VLMs)~\citep{Qwen37Plus, Claude4, gpt5, comanici2025gemini}.

% Human readers rarely process long documents in a single pass.
% Instead, they follow a \textbf{coarse-to-fine reading process}: they first skim or scan the document to build a global view, then selectively inspect relevant pages or regions for fine-grained visual or numerical reasoning, and finally reread the collected evidence to verify the conclusion~\citep{pirolli1999information,keshav2007read,rayner1998eye}.
% This suggests that long-document QA is better modeled as an iterative process of \emph{seeking, reasoning, and verification}, rather than one-shot document-to-answer generation.

% Motivated by this observation, we identify two fundamental capabilities: \textbf{evidence seeking} and \textbf{multimodal understanding}.
% Evidence seeking enables the model to identify what evidence is needed, locate it across documents, pages, regions, and modalities, and continue searching when the current evidence is insufficient. 
% Multimodal understanding enables the model to interpret and connect heterogeneous document elements and reason over their relationships to support the answer.
%
\vspace{-.2em}
Human readers typically follow a \textbf{coarse-to-fine reading process}: they skim for a global view, selectively inspect relevant pages or regions, and revisit evidence to verify their conclusions~\citep{pirolli1999information,keshav2007read,rayner1998eye}.
This motivates modeling long-document QA as an iterative process supported by two core capabilities: \textbf{evidence seeking} and \textbf{multimodal understanding}.
Evidence seeking identifies information needs, locates relevant evidence across documents, pages, and regions, and guides further search when evidence is insufficient.
Multimodal understanding involves interpreting heterogeneous document elements and reasoning over their relationships.

\vspace{-.2em}
Existing approaches remain limited along these two dimensions.
For multimodal understanding, \textbf{OCR-based pipelines} convert documents into textual or structural representations~\citep{xu2020layoutlm}, potentially losing visual cues and propagating parsing errors.
\textbf{Long-context VLMs} directly process page-image sequences~\citep{DBLP:journals/pami/LiuYLLMZB26, DBLP:conf/nips/MaZC0JLLLMDZP0W24}, but face high computational costs, context limits, and the \emph{lost-in-the-middle} effect.
\textbf{Retrieval-augmented generation (RAG)} targets evidence seeking by retrieving relevant pages or passages~\citep{DBLP:conf/emnlp/Saad-FalconBSN024, DBLP:journals/corr/abs-2411-04952, DBLP:conf/iclr/FaysseSWOVHC25}, but conventional static retrieval may overlook evidence dispersed across pages and modalities and lacks the flexibility to adapt to evolving information needs during reasoning.

\begin{figure}[t!]
    \centering
    \includegraphics[width=\linewidth]{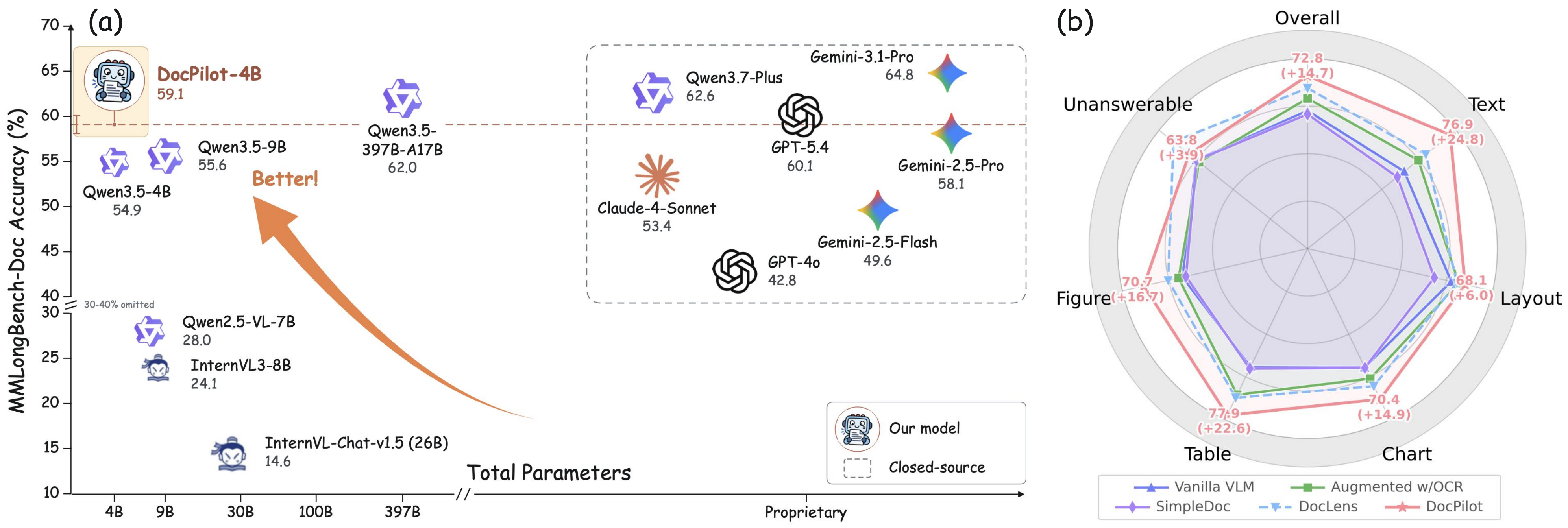}
    \vspace{-2em}
    \caption{ Performance of our proposed \sys on MMLongBench-Doc: (a) competitive accuracy with only 4B parameters; (b) performance gains with Gemini-2.5-Pro as the backbone.
    }
    \label{fig:performance_overview}
    \vspace{-2em}
\end{figure}

\vspace{-.2em}
Recent \textbf{agentic document frameworks} extend static retrieval with multi-step document interactions, but remain limited in two ways.
First, predefined workflows~\citep{DBLP:journals/corr/abs-2503-13964, DBLP:conf/emnlp/JainWZLDSWW25, DBLP:conf/acl/ZhuMCLPY26} limit the flexibility to adapt reasoning paths to intermediate observations.
Second, some agentic systems mainly rely on coarse-grained actions: DocDancer~\citep{DBLP:journals/corr/abs-2601-05163} models document interaction through \textsc{Search} and \textsc{Read}, while Doc-V*~\citep{DBLP:conf/acl/ZhengFLWZRZWLLC26} adopts an OCR-free VLM-based paradigm centered on page-level retrieval and reading.
As illustrated in Fig.~\ref{fig:teaser}, complex DocVQA requires \emph{adaptive evidence seeking} to locate the map and supporting evidence across pages, and \emph{fine-grained multimodal understanding} to identify the leading state, compute the reduction in cost per GW, and check the efficiency criterion.
These coupled requirements motivate our agent orchestration framework, \textbf{\sys}, whose flexible composition of these operations yields substantial gains on MMLongBench-Doc~\citep{DBLP:conf/nips/MaZC0JLLLMDZP0W24}, as shown in Fig.~\ref{fig:performance_overview}.

\vspace{-.2em}
Despite their strong reasoning capabilities, LLM-based document agents can be costly and privacy-sensitive in practical deployment, motivating small language models (SLMs) for efficient, on-premise inference~\citep{liang2026long, li2026dataspace}. 
%
% However, coarse-grained interaction creates a learning mismatch for compact models. 
% Simple trajectories such as \textsc{Search}$\rightarrow$\textsc{Read} are easy to imitate, while each high-level action may implicitly require substantial multimodal reasoning~\citep{DBLP:journals/corr/abs-2601-05163}. 
% As a result, an SLM can learn the interaction pattern without reliably acquiring the capability needed to execute each step. 
% Finer-grained actions make these capabilities explicit and easier to supervise, but make their composition more challenging.
However, training compact models to master the complex orchestration required for document reasoning remains challenging.
Existing approaches based on coarse-grained interactions~\citep{DBLP:journals/corr/abs-2601-05163},
simplify the interaction structure but leave substantial multimodal reasoning implicit within each action.
Consequently, SLMs may imitate action sequences without mastering the capabilities required for reliable execution.
Finer-grained actions facilitate explicit supervision of these capabilities, but increase the complexity of adaptive orchestration.
Reconciling these demands in SLM-based document agents motivates our objective:

\vspace{-.2em}
\emph{A agentic document framework should not only support expressive orchestration, but also turn the complex document-solving process into a learnable objective for SLMs.}

To this end, we introduce \textbf{\sys}, a trainable interactive document agent\footnote{An agent that can be deployed as a trainable model to interact with document environments.} for multimodal long-document QA.
We model heterogeneous document content as a unified multimodal environment, over which \sys orchestrates three complementary capabilities:
(i) adaptive evidence seeking over page-level visual cues and OCR-derived structured representations, guided by intermediate state;
(ii) fine-grained multimodal understanding, supported by specialized operations for local visual inspection, and code-based numerical reasoning; and
(iii) iterative refinement via answer-grounding verification, enabling self-correction.
%
% 我们也解决了reward sparsity and trajectory scarcity
As illustrated in Fig.~\ref{fig:performance_overview} (a), with only 4B parameters, \sys achieves performance competitive with much larger proprietary models.
This performance is enabled by a curriculum-based agentic training paradigm that mitigates \emph{reward sparsity} through progressive capability learning.
The curriculum proceeds in two stages: single-ability supervised fine-tuning (SFT) establishes core document interaction capabilities, while multi-ability agentic training learns to adaptively compose them according to intermediate observations.
To address \emph{trajectory scarcity}, we develop a data-grounded synthesis pipeline that provides high-quality, executable interaction trajectories for this training process.
In summary, our key contributions are three-fold.

\begin{enumerate}[leftmargin=1.5em,itemsep=-1pt,topsep=2pt]
\item \textbf{Agentic Orchestration.}
We introduce\textbf{\sys}, a comprehensive agent orchestration framework for long visual document QA, modeling heterogeneous documents as a unified environment for adaptive evidence seeking, fine-grained multimodal understanding, and iterative refinement.

\item \textbf{Agentic Training.}
We develop a curriculum-based agentic training paradigm with data-grounded trajectory synthesis, enabling compact models to progressively acquire and orchestrate document-solving capabilities through single-ability fine-tuning and multi-ability orchestration learning.

\item \textbf{Empirical Validation.}
Experiments on MMLongBench-Doc, DocBench, FinRAGBench-V, and MADQA demonstrate substantial gains across diverse baselines, while \textbf{\sys-4B} achieves competitive performance against substantially larger proprietary models with only \textbf{4B} parameters.
Analyses such as evidence localization and training ablations, further substantiate robustness.

\end{enumerate}

\section{\sys}
\label{sec:method}
\vspace{-.5em}

\begin{figure}[h!]
    \centering
    \includegraphics[width=\linewidth]{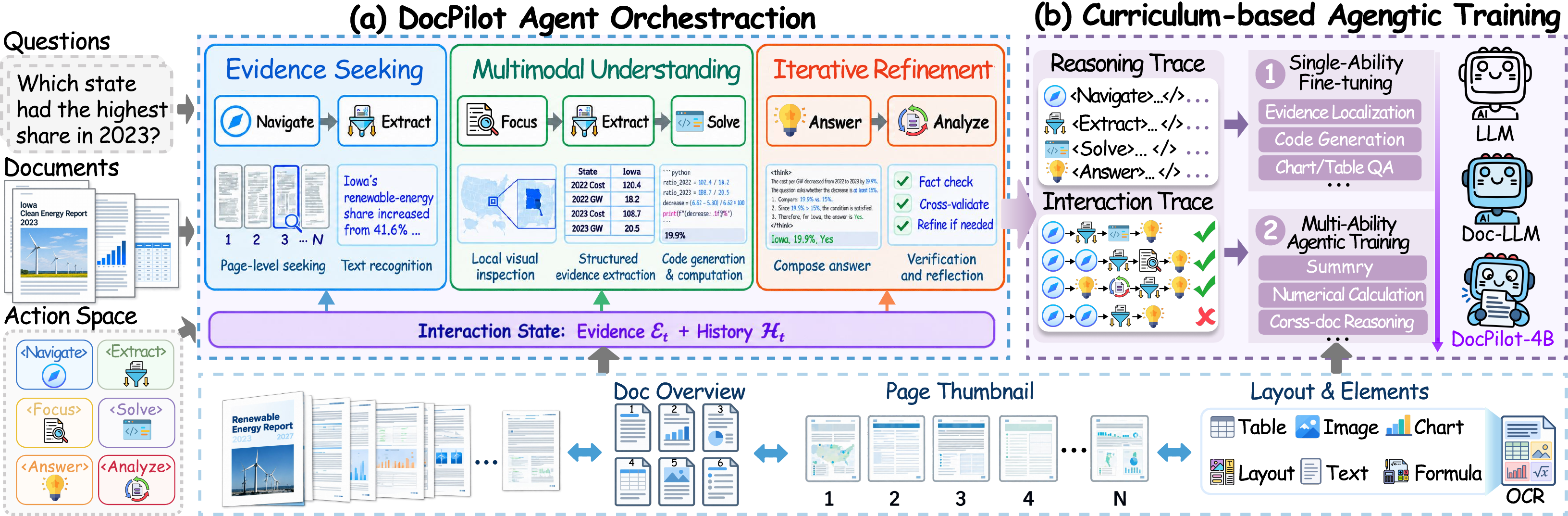}
    \vspace{-2em}
    \caption{
        Overview of \sys.
    }
    \label{fig:overview}
    \vspace{-1em}
 \end{figure}

Fig.~\ref{fig:overview} outlines \sys, a trainable interactive document agent for multimodal long-document QA with two core components.
The \textbf{agent orchestration framework} dynamically selects and composes actions for adaptive evidence seeking, multimodal understanding, and iterative refinement based on intermediate observations.
The \textbf{curriculum-based agentic training} paradigm transfers these capabilities to compact models using high-quality synthesized reasoning and interaction traces, progressing from single-ability fine-tuning to multi-ability agentic training.
Both components share a \textbf{multimodal document environment} for interacting with heterogeneous evidence.

%  Overview部分
%  定义 DocPILOT 是什么；
%  点出两大组件；
%  说明与 multimodal document environment 的交互关系。

\vspace{-.5em}
\subsection{Multimodal Document Environment}
\label{ssec:environment}
\vspace{-.5em}

Given a document $\mathcal{D}=\{I_1,\ldots,I_N\}$, \sys constructs a multimodal environment
$\mathcal{E}(\mathcal{D})=(\{(I_i,\mathcal{S}_i,T_i)\}_{i=1}^{N},\widetilde{\mathcal{D}})$,
where $I_i$ is the original page image, $\mathcal{S}_i$ is the OCR-derived structured representation produced by MinerU2.5~\citep{niu2026mineru2}, $T_i$ is a lightweight page thumbnail, and $\widetilde{\mathcal{D}}$ is the document-level overview.
$\mathcal{S}_i$ preserves recognized content, layout elements, reading order, and spatial locations, while $T_i$ is a low-resolution page thumbnail (\eg 512$\times$512) that retains coarse layout and salient visual cues with its page number explicitly marked.
The document overview further arranges consecutive thumbnails into grids to provide a compact global view, with separate overviews constructed for multi-document inputs.
All representations share consistent document and page identifiers, enabling aligned access to textual, structural, and visual content.

\subsection{Agent Orchestration Framework}
\label{sec:agent_orchestration}

\noindent\textbf{Problem Formulation.}
We formulate multimodal long-document QA as an evidence-driven interaction process between the agent and a document environment.
Rather than following a predefined sequence of operations or repeated sampling, \sys continuously updates its evidence state and selects the next action according to the unresolved information need.
This yields a closed-loop process of \emph{seeking, reasoning, verification, and refinement}, as summarized in Algorithm~\ref{alg:docpilot_inference}.
\vspace{-.5em}

% \begin{minipage}[t]{0.52\textwidth}
%     \begin{algorithm}[H]
%         \footnotesize
%         \renewcommand{\baselinestretch}{0.88}\selectfont
%         \SetAlCapSkip{2pt}
%         \SetAlgoInsideSkip{2pt}

%         \caption{Inference of \sys}
%         \label{alg:docpilot_inference}

%         \KwIn{Question $q$, environment $\mathcal{D}$, model $\mathcal{M}$}
%         \KwOut{Grounded answer $y$}

%         Initialize evidence $\mathcal{E}\leftarrow\emptyset$ and
%         history $\mathcal{H}\leftarrow\emptyset$\;

%         \While{answer not verified}{
%             $(a,\phi)\leftarrow\mathcal{M}(q,\mathcal{E},\mathcal{H})$
%             \tcp*{select next action}

%             \uIf{$a\in\{\textsc{Navigate},\textsc{Extract},
%             \textsc{Focus},\textsc{Solve}\}$}{
%                 $o\leftarrow\mathcal{D}.\textsc{Execute}(a,\phi)$\;
%                 $\mathcal{E}\leftarrow\textsc{Update}(\mathcal{E},o)$\;
%                 $\mathcal{H}\leftarrow\mathcal{H}\cup\{(a,o)\}$\;
%             }
%             \ElseIf{$a=\textsc{Answer}$}{
%                 $\hat{y}\leftarrow\phi$\;
%                 $(v,f)\leftarrow
%                 \mathcal{M}.\textsc{Analyze}(q,\mathcal{E},\hat{y})$\;

%                 \uIf{$v=\texttt{supported}$}{
%                     \Return{$\hat{y}$}\;
%                 }
%                 \Else{
%                     $\mathcal{E}\leftarrow\textsc{Refine}(\mathcal{E},f)$\;
%                 }
%             }
%         }
%     \end{algorithm}
% \end{minipage}

\newcommand{\algactiontag}[2]{%
    \tcbox[
        on line,
        colback=#1!15,
        colframe=#1,
        boxrule=0.35pt,
        arc=1pt,
        boxsep=0.6pt,
        left=1.8pt,
        right=1.8pt,
        top=0.6pt,
        bottom=0.6pt
    ]{\scriptsize\textsf{#2}}%
}

\noindent
\begin{minipage}[t]{0.46\textwidth}
    \vspace{0pt}
    \small
    \renewcommand{\baselinestretch}{0.92}\selectfont

    \setlength{\abovedisplayskip}{2pt}
    \setlength{\belowdisplayskip}{2pt}
    \setlength{\abovedisplayshortskip}{1pt}
    \setlength{\belowdisplayshortskip}{1pt}
    
    Given a question $q$ and a multimodal document environment $\mathcal{D}$, a document agent typically seeks relevant evidence, interprets multimodal content, and iteratively refines its reasoning until a grounded answer is obtained. To support this process, \sys introduces six specialized actions, including:
    \begin{itemize}[leftmargin=1.2em, itemsep=2pt, topsep=2pt]

        \item \actiontag{navigatecolor}{Navigate}:
        Locate relevant visual pages.
    
        \item \actiontag{extractcolor}{Extract}:
        Extract task-relevant evidence from selected pages or regions.
    
        \item \actiontag{focuscolor}{Focus}:
        Inspect local visual regions through cropping, zooming, or subfigure focusing.
    
        \item \actiontag{solvecolor}{Solve}:
        Perform code-based numerical reasoning over collected evidence.
    
        \item \actiontag{analyzecolor}{Analyze}:
        Assess answer grounding for planning, verification, and refinement.
    
        \item \actiontag{answercolor}{Answer}:
        Produce a candidate answer.
    
    \end{itemize}
    Failed verification triggers further evidence seeking or reasoning, closing the loop through three complementary mechanisms: adaptive evidence seeking, multimodal understanding, and iterative refinement.

\end{minipage}
\hfill
\begin{minipage}[t]{0.48\textwidth}
    \vspace{0pt}
    \begin{algorithm}[H]
        \footnotesize
        \renewcommand{\baselinestretch}{0.80}\selectfont
        \SetAlCapSkip{2pt}
        \SetAlgoInsideSkip{2pt}

        \caption{Inference of \sys}
        \label{alg:docpilot_inference}

        \KwIn{Question $q$, environment $\mathcal{D}$, model $\mathcal{M}$}
        \KwOut{Grounded answer $y$}

        Initialize evidence $\mathcal{E}\leftarrow\emptyset$ and
        history $\mathcal{H}\leftarrow\emptyset$\;

        \While{answer not verified}{
            $(a,\phi)\leftarrow\mathcal{M}(q,\mathcal{E},\mathcal{H})$
            \hspace{0.3em}{\itshape // select next action}\;

            \uIf{$a\in\{
            \mbox{\algactiontag{navigatecolor}{Navigate}},
            \mbox{\algactiontag{extractcolor}{Extract}},
            \mbox{\algactiontag{focuscolor}{Focus}},
            \mbox{\algactiontag{solvecolor}{Solve}}
            \}$}{
                $o\leftarrow\mathcal{D}.\textsc{Execute}(a,\phi)$\;
                $\mathcal{E}\leftarrow\textsc{Update}(\mathcal{E},o)$\;
                $\mathcal{H}\leftarrow\mathcal{H}\cup\{(a,o)\}$\;
            }
            \ElseIf{$a=\mbox{\actiontag{answercolor}{Answer}}$}{
                $\hat{y}\leftarrow\phi$\;
                $(v,f)\leftarrow
                \mathcal{M}.\mbox{\actiontag{analyzecolor}{Analyze}}
                (q,\mathcal{E},\hat{y})$\;

                \uIf{$v=\texttt{supported}$}{
                    \Return{$\hat{y}$}\;
                }
                \Else{
                    $\mathcal{E}\leftarrow
                    \textsc{Refine}(\mathcal{E},f)$\;
                }
            }
        }
    \end{algorithm}
\end{minipage}

% \vspace{0.5em}

\noindent\textbf{Adaptive Evidence Seeking.}
\sys combines \texttt{<Navigate>} and \texttt{<Extract>} to seek evidence at complementary granularities: \texttt{<Navigate>} scans page thumbnails to identify potentially relevant pages while preserving broad coverage with low visual-token cost, and \texttt{<Extract>} further operates on OCR-derived text, tables, and parsed visual elements to obtain structured, source-grounded evidence:
\begin{equation}
    P_t = \textsc{Navigate}(q_t,\mathcal{E}_t,\{T_i\}),
    \qquad
    e_t = \textsc{Extract}(q_t,\mathcal{E}_t,P_t).
\end{equation}
where $\{T_i\}$ denotes the page thumbnails, $P_t$ the selected pages, and $e_t$ the extracted structured evidence. If the evidence remains insufficient, subsequent search targets are updated from the evolving evidence state, enabling state-driven, coarse-to-fine evidence recovery.

\noindent\textbf{Multimodal Understanding.}
To mitigate the errors introduced by directly reasoning over heterogeneous content, \texttt{<Focus>} and \texttt{<Solve>} serve distinct yet coordinated roles:
\begin{equation}
    z_t = \textsc{Focus}(e_t;\omega_t),
    \qquad
    c_t = \textsc{Solve}(e_t),\;
    r_t = \textsc{ExecuteCode}(c_t).
\end{equation}
\textsc{Focus} selectively applies a visual operation $\omega_t \in \{\mathrm{Crop}, \mathrm{Zoom}, \mathrm{Isolate}, \ldots\}$ to the relevant evidence $e_t$ when finer visual inspection is required, yielding the refined observation $z_t$. 
Meanwhile, \textsc{Solve} generates executable code $c_t$ for aggregation, comparison, or multi-step numerical reasoning, whose execution produces result $r_t$ and updates the evidence state. Together, these actions enable precise visual inspection and reliable numerical reasoning over multimodal evidence.

\noindent\textbf{Iterative Refinement.}
Once sufficient evidence has been accumulated, \texttt{<Answer>} synthesizes a candidate response, which is subsequently examined by \texttt{<Analyze>} for adequacy and faithfulness:
\begin{equation}
    \hat{y}_t = \textsc{Answer}(q,\mathcal{E}_t),
    \qquad
    (v_t,f_t) = \textsc{Analyze}(q,\mathcal{E}_t,\hat{y}_t),
\end{equation}
where $v_t$ denotes the verification outcome and $f_t$ the feedback on remaining evidence gaps or inconsistencies. A fully supported candidate is accepted; otherwise, $f_t$ is fed back into the interaction loop to resume evidence seeking and multimodal reasoning, turning verification into an explicit mechanism for iterative self-correction.
Further construction details of \sys, including the document environment, interaction protocol, and individual module designs, are provided in Appendix~\ref{app:construction}.

\subsection{Curriculum-based Agentic Training}
\label{sec:curriculum_based_agentic_training}

% \begin{figure}[h!]
%     \centering
%     \includegraphics[width=\linewidth]{fig/curriculum_training.pdf}
%     \vspace{-1em}
%     \caption{
%         Curriculum-based Training of \sys~\lzw{need to update}.
%     }
%     \label{fig:overview}
%     \vspace{-1em}
%  \end{figure}

\noindent\textbf{Data-grounded Trajectory Synthesis.}
Existing DocVQA data largely provide only question--answer supervision, leaving both action-level reasoning and cross-action orchestration under-specified. To support our curriculum training, we therefore synthesize two complementary forms of supervision: \textit{reasoning traces} for learning how to execute individual capabilities, and \textit{interaction traces} for learning when and how to compose them according to intermediate observations. Together, these traces form \textbf{\textsc{DocAgent-Instruct-40K}}, including 35K single-ability reasoning traces and 5K multi-ability interaction traces for agentic training. More details are provided in Appendix~\ref{sec:app_synthesis}.

\textit{Reasoning Trace Synthesis.}
For single-ability supervision, we perform teacher-guided reasoning distillation for page-level navigation, element-level extraction, and reasoning (\eg code generation, table QA, and chart QA). We draw evidence-seeking instances from DocSeeker~\citep{yan2026docseeker} and HotpotQA~\citep{yang2018hotpotqa}, and reasoning instances from FinQA~\citep{chen2021finqa}, ChartQAPro~\citep{masry2025chartqapro}, and InfographicVQA~\citep{mathew2022infographicvqa}. A strong teacher distills capability-specific traces from evidence and target answers, filtered by ground-truth consistency.

\textit{Interaction Trace Synthesis.}
For multi-ability supervision, we draw QA instances from LongDocURL~\citep{deng2025longdocurl}, DUDE~\citep{van2023document}, MADQA~\citep{borchmann2026strategic}, and MMDocRAG~\citep{dong2026benchmarking}, and roll out a strong teacher agent in the \sys environment. Conditioned on the question, interaction history, and evolving evidence state, the teacher iteratively executes actions to produce a grounded answer. These trajectories capture evidence seeking, multimodal reasoning, and verification. We retain only trajectories with correct answers, valid interactions, and well-grounded evidence, filtering out failed or redundant executions.

\begin{wrapfigure}{r}{0.52\linewidth}
    \vspace{-\intextsep} % 抵消顶部默认间距
    \centering
    \includegraphics[width=\linewidth]{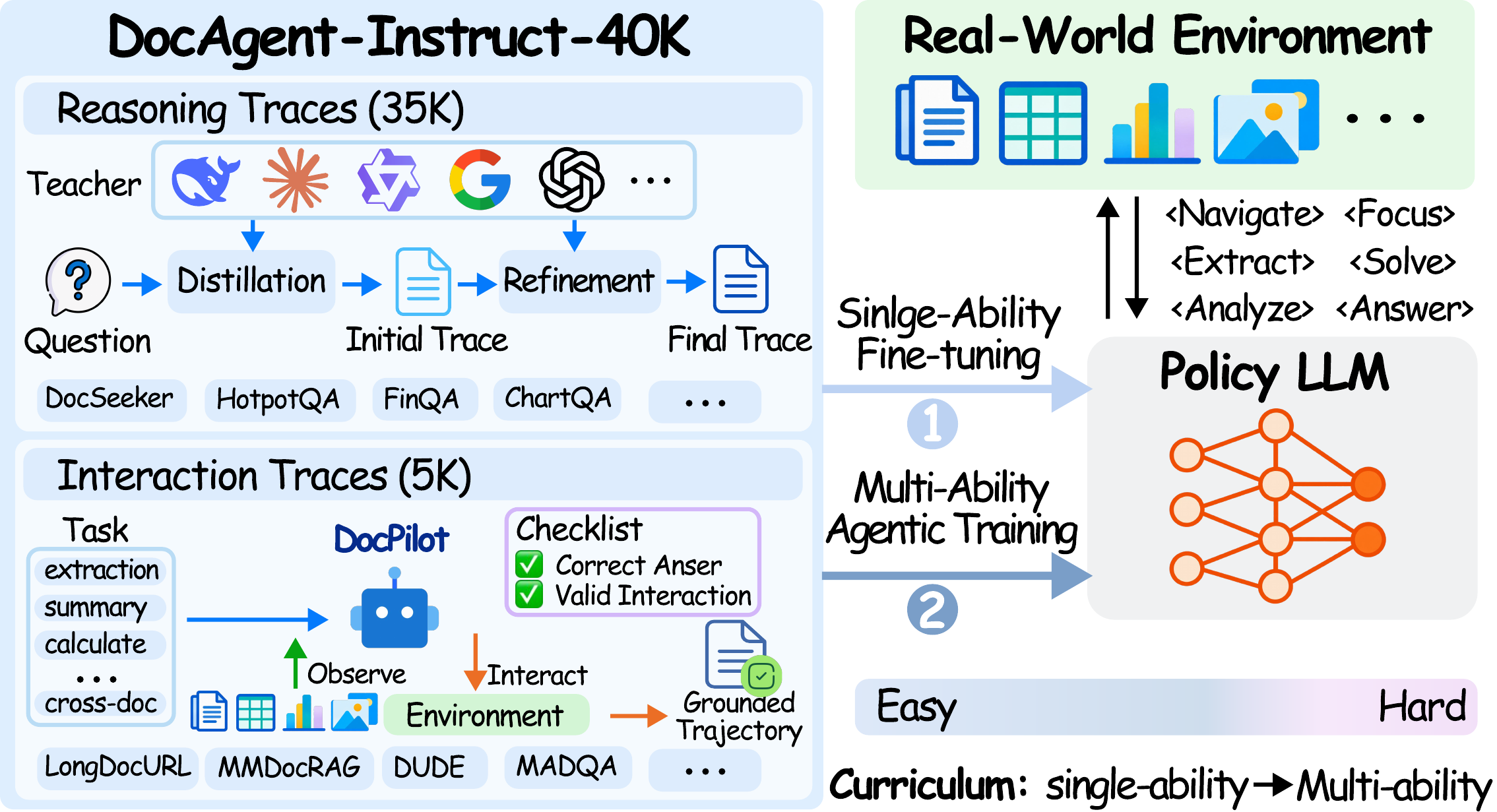}
    \vspace{-2em}
    \caption{Curriculum-based training.}
    \label{fig:curriculum_training}
    \vspace{-2em}
\end{wrapfigure}
\noindent\textbf{Single-Ability Fine-tuning.}
Since foundation LLMs are not explicitly trained for interactive document solving, we first strengthen the specialized capabilities required by \sys. As illustrated in Fig.~\ref{fig:curriculum_training}, this stage covers page-level navigation (\texttt{<Navigate>}), element-level evidence extraction (\texttt{<Extract>}), code-based numerical reasoning (\texttt{<Solve>}), and multimodal question answering (\texttt{<Answer>}). We fine-tune the model on teacher-generated reasoning traces using next-token prediction to support subsequent orchestration learning.

\noindent\textbf{Multi-Ability Agentic Training.}
Building on the acquired single abilities, we further train the model on synthesized multi-ability interaction trajectories to learn adaptive orchestration for complex document solving. Each trajectory interleaves agent reasoning and actions with environment observations, culminating in a grounded answer. This supervision extends beyond isolated capability execution to tool selection, action composition, and strategy adjustment in response to intermediate evidence.

We optimize a masked autoregressive objective, retaining environment observations as conditioning context while supervising only agent-generated outputs. Given the task context \(c\) and tokenized trajectory \(H=(x_1,\ldots,x_L)\), the training objective is:
\begin{equation}
\mathcal{L}_{\mathrm{multi}}(\theta)
=
-\mathbb{E}_{(c,H)\sim\mathcal{D}_{\mathrm{multi}}}
\left[
\frac{1}{\sum_{t=1}^{L}m_t}
\sum_{t=1}^{L}
m_t\log\pi_\theta(x_t\mid c,x_{<t})
\right].
\label{eq:multi_ability_training}
\end{equation}
where \(\mathcal{D}_{\mathrm{multi}}\) denotes the interaction-trajectory training set, and \(m_t\) equals one for agent-generated reasoning, actions, and answers, and zero otherwise. Environment feedback thus informs subsequent decisions without serving as a prediction target.

\section{Experiments}
\label{sec:experiment}

We conduct experiments to answer the following research questions:

\textbf{RQ1:} How effective are \sys's orchestration framework and training strategy on multimodal long-document QA?
\textbf{RQ2:} How much does each core component of \sys contribute to the final performance?
\textbf{RQ3:} Does \sys provide stronger evidence seeking than existing retrieval or agentic document systems?
\textbf{RQ4:} How does \sys behave during interaction, and can it generalize to more challenging multi-document scenarios?

\subsection{Experimental Setup}
\label{sec:exp_setup}

\noindent \textbf{Benchmarks and Datasets.}
We evaluate \sys on two challenging general-purpose multimodal long-document QA benchmarks: MMLongBench-Doc~\citep{DBLP:conf/nips/MaZC0JLLLMDZP0W24} and DocBench~\citep{zou2025docbench}. 
MMLongBench-Doc contains 1,082 questions over 135 long PDFs across seven domains, averaging 47.5 pages with rich multimodal layouts, while DocBench includes 1,102 questions over 229 real-world documents spanning five domains and four major question types.
For domain-specific evaluation, we use FinRAGBench-V~\citep{zhao2025finragbench}, which targets realistic financial document understanding and provides fine-grained evidence annotations for evaluating our seeking strategy.
Cross-document generalization is evaluated on MADQA~\citep{borchmann2026strategic}, a benchmark grounded in heterogeneous document collections and designed for agentic reasoning.

\noindent \textbf{Metrics.}
We follow each benchmark's official evaluation protocol.
For MMLongBench-Doc, we report rule-based accuracy and F1, while FinRAGBench-V and DocBench use LLM-as-a-judge (\textit{LasJ}) evaluation with GPT-4o.
For MADQA, we report overall, cross-page (\textit{X-Page}), and cross-document (\textit{X-Doc}) accuracy, with Page F1 for evidence attribution and Kuiper statistic for effort calibration.

\noindent \textbf{Models and Baselines.}
We evaluate \sys on three baselines categories: (1) vanilla VLMs (Qwen3.5-4B/9B/397B-A17B, GPT-4o, GPT-5.4~\citep{gpt5}, Qwen3.7-Plus~\citep{Qwen37Plus}, Claude-4-Sonnet~\citep{Claude4}, and Gemini-2.5/3.1 models);
(2) OCR-based methods using Claude-4-Sonnet and Gemini-2.5-Flash/Pro~\citep{comanici2025gemini}; and (3) Agentic frameworks including M3DocRAG~\citep{DBLP:journals/corr/abs-2411-04952}, MDocAgent~\citep{DBLP:journals/corr/abs-2503-13964}, DocDancer~\citep{DBLP:journals/corr/abs-2601-05163}, SimpleDoc~\citep{DBLP:conf/emnlp/JainWZLDSWW25}, and DocLens~\citep{DBLP:conf/acl/ZhuMCLPY26}. 
On MADQA, we additionally compare with OpenAI File Search, HEAVEN~\citep{kim2026hybrid}, Codex, BM25 Agent, and Recursive Language Models (RLM)~\citep{zhang2025recursive}, all with GPT-5.4 as the backbone.

\noindent \textbf{Implementation Details.} 
\sys adopt the vanilla ReAct~\citep{yao2022react} as the agent's framework.
% During the curriculum-based training, \sys employs a two-phase training pipeline comprising LoRA fine-tuning followed by GRPO optimization using ms-swift and verl.
% During single-ability SFT, we train the model for 3 epochs with a learning rate of $10^{-4}$, a batch size of 64, and a LoRA adapter with rank $r=16$ and $\alpha=32$.
% During GRPO, we use a learning rate of $10^{-6}$, sample 6 generations per query, and apply a 25k-token generation limit (also used for multi-ability cold-start) to accommodate long interaction trajectories.
% More details are provided in Appendix~\ref{}.
We implement curriculum-based training in ms-swift\footnote{\url{https://github.com/modelscope/ms-swift}}, comprising single-ability SFT followed by multi-ability agentic training.
Both stages use 3 training epochs, a learning rate of $10^{-4}$, a batch size of 64, and LoRA adapters with rank $r=16$ and $\alpha=32$.
The maximum sequence lengths are 8K and 24K text tokens for the two stages, respectively, excluding visual tokens.

\subsection{Main Results}
\label{sec:main_results}

\begin{table*}[t]
    \centering
    \small
    \caption{Performance comparison on multimodal long-document QA benchmarks. Results are reported by evidence source: text (TXT), layout (LAY), chart (CHA), table (TAB), figure (FIG), and unanswerable (UNA), with overall accuracy and F1. $\dagger$ denotes the reused result from prior papers. 
    \colorbox{firstBest}{Blue} highlights the best, while \colorbox{secondBest}{Red} indicates the second-best.}
    \vspace{-1em}
    \label{tab:main_results}
    \setlength{\tabcolsep}{3.0pt}
    \renewcommand{\arraystretch}{1.05}
    
    \resizebox{\linewidth}{!}{
    \begin{tabular}{l cccccccc|cccc|c}
    \toprule
    \multirow{2}{*}{\textbf{Model}}
    & \multicolumn{8}{c|}{\textbf{MMLongBench-Doc}}
    & \multicolumn{4}{c|}{\textbf{FinRAGBench-V}}
    & \multicolumn{1}{c}{\textbf{DocBench}} \\
    \cmidrule(lr){2-9}
    \cmidrule(lr){10-13}
    \cmidrule(lr){14-14}
    & \textbf{TXT}
    & \textbf{LAY}
    & \textbf{CHA}
    & \textbf{TAB}
    & \textbf{FIG}
    & \textbf{UNA}
    & \textbf{Acc}
    & \textbf{F1}
    & \textbf{TXT}
    & \textbf{TAB}
    & \textbf{CHA}
    & \textbf{LasJ}
    & \textbf{LasJ} \\
    \midrule
    
    % ============================================================
    % Vanilla VLMs
    % ============================================================
    \multicolumn{14}{c}{\textit{Vanilla VLMs}} \\
    \midrule
    
    \multicolumn{14}{l}{\textcolor{gray}{\textit{Open-source Models}}} \\
    
    Qwen3.5-4B
    & 54.6 & 57.7 & 50.8 & 57.1 & 47.2 & 54.7 & 54.9 & 53.5
    & 42.4 & 50.4 & 65.8 & 52.7
    & 69.3 \\
    
    Qwen3.5-9B
    & 55.6 & 56.3 & 47.5 & 52.5 & 47.9 & 60.1 & 55.6 & 55.2
    & 43.1 & 55.5 & 58.1 & 52.9
    & 69.8 \\
    
    Qwen3.5-397B-A17B
    & 65.1 & 73.9 & 66.1 & 72.3 & 60.8 & 36.8 & 62.0 & 63.3
    & 52.1 & 64.7 & 69.0 & 62.3
    & 80.6 \\
    
    \multicolumn{14}{l}{\textcolor{gray}{\textit{Proprietary Models}}} \\
    
    GPT-4o
    & 46.3 & 46.0 & 45.3 & 50.0 & 44.1 & 20.2 & 42.8 & 44.9
    & 53.9 & 33.3 & 52.6 & 43.4
    & 63.1 \\
    
    GPT-5.4
    & 62.3 & 62.2 & 65.0 & 75.2 & 58.6 & 31.8 & 60.1 & 61.8
    & 59.0 & 49.3 & 60.0 & 55.1
    & 73.5 \\
    
    Qwen3.7-Plus
    & 67.7 & 76.5 & 64.4 & 75.5 & 69.0 & 26.9 & 62.6 & 63.7
    & 55.6 & 69.8 & 69.7 & 65.9
    & 83.2 \\
    
    Claude-4-Sonnet
    & 50.4 & 49.4 & 50.5 & 57.3 & 43.9 & 59.0 & 53.4 & --
    & 36.6 & 20.2 & 51.9 & 33.8
    & -- \\
    
    Gemini-2.5-Flash
    & 44.0 & 53.2 & 46.0 & 43.9 & 48.2 & 56.7 & 49.6 & --
    & 49.0 & 41.6 & 41.0 & 43.0
    & -- \\
    
    Gemini-2.5-Pro
    & 52.1 & 62.1 & 55.5 & 55.3 & 54.0 & 59.9 & 58.1 & --
    & 62.2 & 55.3 & 50.4 & 54.9
    & 74.8 \\
    
    Gemini-3.1-Pro
    & 68.0 & 73.9 & 66.1 & 74.3 & 65.5 & 39.9 & 64.8 & 66.6
    & 50.0 & 72.7 & 64.5 & 64.3
    & 80.6 \\
    
    \midrule
    
    % ============================================================
    % VLMs Augmented with OCR
    % ============================================================
    \multicolumn{14}{c}{\textit{VLMs Augmented with OCR}} \\
    \midrule
    
    Claude-4-Sonnet
    & 52.7 & 51.6 & 50.0 & 58.1 & 45.3 & 65.9 & 56.0 & --
    & 58.7 & 21.6 & 54.3 & 41.0
    & -- \\
    
    Gemini-2.5-Flash
    & 55.9 & 54.9 & 52.7 & 63.4 & 50.3 & 60.8 & 58.5 & --
    & 67.6 & 64.4 & 46.1 & 58.3
    & -- \\
    
    Gemini-2.5-Pro
    & 59.7 & 65.3 & 60.8 & 68.3 & 55.7 & 58.4 & 63.3 & --
    & 70.0 & 70.0 & 56.2 & 64.9
    & -- \\
    
    \midrule
    
    % ============================================================
    % VLM-based Agentic Frameworks
    % ============================================================
    \multicolumn{14}{c}{\textit{VLM-based Agentic Frameworks}} \\
    \midrule
    
    M3DocRAG (w/ Qwen2-VL-7B)$^{\dagger}$
    & 30.0 & 23.5 & 18.9 & 20.1 & 20.8 & 5.8 & 21.0 & --
    & -- & -- & -- & --
    & -- \\
    
    MDocAgent (w/ GPT-4o)$^{\dagger}$
    & -- & -- & -- & -- & -- & -- & 42.0 & --
    & -- & -- & -- & --
    & -- \\
    
    DocDancer (FT w/ Qwen3-4B)$^{\dagger}$
    & -- & -- & -- & -- & -- & -- & 48.4 & 49.2
    & -- & -- & -- & --
    & 79.8 \\
    
    DocDancer (w/ Gemini-2.5-Pro)$^{\dagger}$
    & -- & -- & -- & -- & -- & -- & 56.3 & 55.3
    & -- & -- & -- & --
    & 79.9 \\
    
    \textbf{SimpleDoc}
    & & & & & & & & & & & & & \\
    
    \quad w/ Claude-4-Sonnet
    & 52.1 & 53.3 & 58.3 & 62.4 & 46.9 & 66.5 & 58.6 & --
    & 59.6 & 68.9 & 54.9 & 61.7
    & -- \\
    
    \quad w/ Gemini-2.5-Flash
    & 45.5 & 57.4 & 49.0 & 51.6 & 45.2 & 66.5 & 53.3 & --
    & 70.2 & 56.2 & 53.6 & 58.3
    & -- \\
    
    \quad w/ Gemini-2.5-Pro
    & 48.4 & 54.8 & 55.7 & 56.1 & 52.5 & 59.7 & 56.6 & --
    & 67.5 & 64.0 & 60.9 & 63.6
    & -- \\
    
    \textbf{DocLens}$^{\dagger}$
    & & & & & & & & & & & & & \\
    
    \quad w/ Claude-4-Sonnet
    & 59.9 & 58.2 & 54.4 & 63.9 & 55.3 & 74.0 & 63.3 & --
    & 70.2 & 66.0 & 60.3 & 64.8
    & -- \\
    
    \quad w/ Gemini-2.5-Flash
    & 59.5 & 61.5 & 54.8 & 66.9 & 59.0 & 73.8 & 64.7 & --
    & 69.9 & 71.3 & 64.5 & 68.5
    & -- \\
    
    \quad w/ Gemini-2.5-Pro
    & 63.7 & 64.6 & 64.3 & 69.7 & 60.2 & 72.2 & 67.6$^{*}$ & --
    & 68.9 & 74.2 & 67.1 & 70.4
    & -- \\
    
    \midrule
    
    % ============================================================
    % DocPILOT
    % ============================================================
    \rowcolor{gray!15}
    \multicolumn{14}{l}{\textbf{\sys (Ours)}} \\
    \midrule
    
    \quad w/ Qwen3.5-397B-A17B
    & 71.8 & 52.1 & 65.5 & 72.5 & 63.3 & \cellcolor{firstBest} 83.0 & 72.1 & 68.8
    & \cellcolor{firstBest} 78.6 & 66.8 & 74.0 & 72.1
    & \cellcolor{secondBest} 84.3\\
    
    \quad w/ Gemini-2.5-Pro
    & \cellcolor{firstBest} 76.9 & 68.1 & \cellcolor{secondBest} 70.4 & \cellcolor{firstBest} 77.9 & 70.7 & 63.8 & 72.8 & \cellcolor{secondBest} 73.0
    & \cellcolor{secondBest} 72.0 & 72.6 & 75.8 & 73.4
    & 84.1 \\
    
    \quad w/ Gemini-3.1-Pro
    & \cellcolor{secondBest} 75.4 & \cellcolor{firstBest} 70.0 & \cellcolor{firstBest} 74.6 & \cellcolor{secondBest} 74.5 & \cellcolor{firstBest} 74.2 &  73.2 & \cellcolor{firstBest} 75.6 & \cellcolor{firstBest} 74.9
    & 68.3 & \cellcolor{firstBest} 79.4 & \cellcolor{secondBest} 77.4 & \cellcolor{firstBest} 75.8
    & \cellcolor{firstBest} 86.8 \\
    
    \quad w/ GPT-5.4
    & 72.7 & \cellcolor{secondBest} 68.4 & 69.1 & 71.6 & \cellcolor{secondBest} 71.4 & \cellcolor{secondBest} 75.0 & \cellcolor{secondBest} 73.5 & 72.4
    & 59.3 & \cellcolor{secondBest} 79.4 & \cellcolor{firstBest} 78.7 & \cellcolor{secondBest} 73.8
    & 83.5 \\

    \midrule

    \quad \sys-4B
    & 57.4\textsubscript{\textcolor{green!50!black}{(+2.8)}}
    & 47.9\textsubscript{\textcolor{red!75!black}{(-9.8)}}
    & 53.4\textsubscript{\textcolor{green!50!black}{(+2.6)}}
    & 58.7\textsubscript{\textcolor{green!50!black}{(+1.6)}}
    & 51.0\textsubscript{\textcolor{green!50!black}{(+3.8)}}
    & 70.0\textsubscript{\textcolor{green!50!black}{(+15.3)}}
    & 59.1\textsubscript{\textcolor{green!50!black}{(+4.2)}}
    & 58.5\textsubscript{\textcolor{green!50!black}{(+5.0)}}
    & 68.7\textsubscript{\textcolor{green!50!black}{(+26.3)}}
    & 56.2\textsubscript{\textcolor{green!50!black}{(+5.8)}}
    & 59.6\textsubscript{\textcolor{red!75!black}{(-6.2)}}
    & 60.2\textsubscript{\textcolor{green!50!black}{(+7.5)}}
    & 80.3\textsubscript{\textcolor{green!50!black}{(+11.0)}} \\
    
    % \quad RL w/ Qwen3.5-9B
    % & -- & -- & -- & -- & -- & -- & -- & --
    % & -- & -- & -- & --
    % & -- \\
    
    \midrule
    
    Human Expert$^{\dagger}$
    & -- & -- & -- & -- & -- & -- & 65.8 & 66.0
    & -- & -- & -- & --
    & -- \\
    
    \bottomrule
    \end{tabular}
    }
\vspace{-1em}
\end{table*}

Table~\ref{tab:main_results} summarizes the main results across MMLongBench-Doc, FinRAGBench-V, and DocBench.

\noindent\textbf{Large gains across LLMs.}
Despite the strong standalone performance of modern VLM backbones, \sys delivers substantial gains across all evaluated models.
On the general-purpose benchmarks, \sys improves vanilla backbones by 10.1, 14.7, 10.8, and 13.4 points on MMLongBench-Doc, and 3.7, 9.3, 6.2, and 10.0 points on DocBench for Qwen3.5-397B-A17B, Gemini-2.5-Pro, Gemini-3.1-Pro, and GPT-5.4, respectively.
On FinRAGBench-V, the corresponding gains reach 9.8, 18.5, 11.5, and 18.7 LasJ points.
These results demonstrate that \sys complements strong base models, with consistent benefits across backbones and benchmarks.

\noindent\textbf{Advantages over existing agentic frameworks.}
\sys also consistently outperforms prior document agents under comparable backbones.
With Gemini-2.5-Pro, \sys improves MMLongBench-Doc accuracy from 67.6 with DocLens to 72.8, and FinRAGBench-V LasJ from 70.4 to 73.4.
On DocBench, it reaches 84.1 LasJ with the same backbone, exceeding DocDancer's 79.9,
highlighting the effectiveness of \sys's richer orchestration over existing agentic pipelines.

\noindent\textbf{Robustness across multimodal capabilities.}
A finer-grained breakdown reveals three complementary capability patterns.
First, semantic and structural understanding improves substantially: TXT gains +24.8 with Gemini-2.5-Pro and +10.4 with GPT-5.4, while LAY increases by +6.0 and +6.2, respectively.
Second, gains are strongest on visually grounded reasoning. Gemini-2.5-Pro raises CHA/TAB/FIG by +14.9/+22.6/+16.7 on MMLongBench-Doc, while GPT-5.4 delivers +24.5/+35.7 on CHA/TAB on FinRAGBench-V.
Finally, answer reliability is markedly strengthened, with Unanswerable (UNA) rising from 31.8 to 75.0 for GPT-5.4 and from 39.9 to 73.2 for Gemini-3.1-Pro.
Overall, \sys exhibits strong and balanced capabilities across semantic understanding, fine-grained multimodal reasoning, and evidence-grounded answerability.

\noindent\textbf{Compact agentic models $\approx$ LLM.}
\sys-4B consistently outperforms vanilla Qwen3.5-4B, with absolute gains of 4.2\%, 7.5\%, and 11.0\% on MMLongBench-Doc, FinRAGBench-V, and DocBench, respectively.
These improvements support the effectiveness of curriculum-based training in developing and coordinating document-solving capabilities.
With only 4B parameters, it surpasses vanilla Gemini-2.5-Pro across all three benchmarks and GPT-5.4 on the latter two, while nearly matching Gemini-3.1-Pro on DocBench (80.3 vs.\ 80.6).
These results highlight learned document interaction as a route to competitive performance at a compact model scale.

\begin{figure}[!b]
    \centering
    \includegraphics[width=\linewidth]{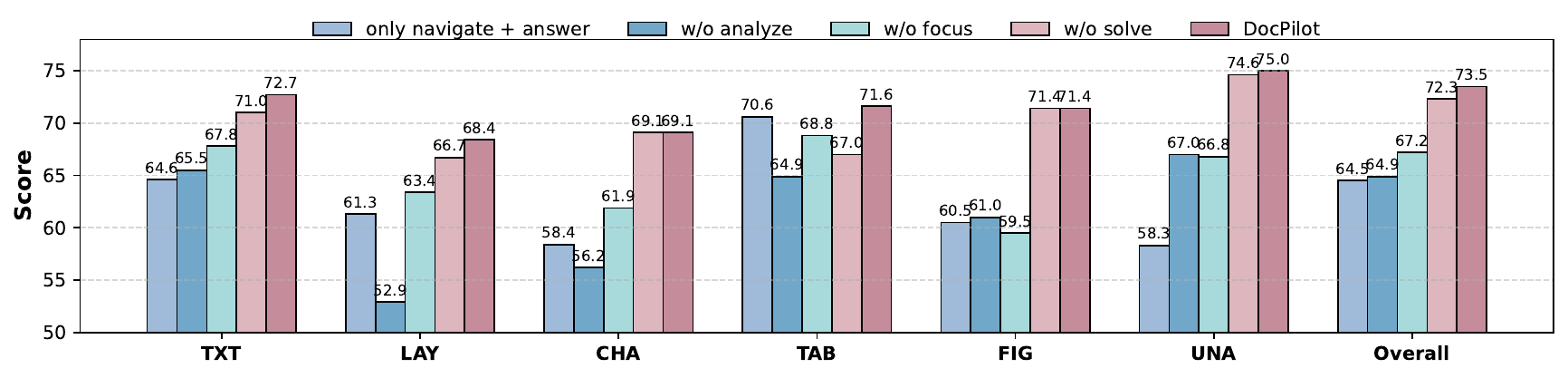}
    \vspace{-2em}
    \caption{
        Ablation of core tools in \sys on MMLongBench-Doc.
    }
    \vspace{-1em}
    \label{fig:tool_ablation}
 \end{figure}

\subsection{Ablation Studies}
\label{sec:ablation}

We conduct ablation studies to understand which components are most important for \sys.
The analysis is divided into two parts: ablation on core actions and ablation on the training strategy.

\noindent \textbf{Ablation on Core Modules.}
% Analyze the effect of removing verification, extract, focus, solve, and using only navigate + answer.
Fig.~\ref{fig:tool_ablation} presents ablations on MMLongBench-Doc using GPT-5.4 as the backbone, highlighting the complementary roles of specialized tools across evidence sources.
Removing \texttt{<Focus>} reduces chart and figure accuracy by 7.2 and 11.9 points, respectively, demonstrating the importance of local visual inspection.
The 4.6\% drop in table accuracy without \texttt{<Solve>} supports code-based numerical reasoning.
Ablating \texttt{<Analyze>} reduces accuracy from 73.5 to 64.9, leaving only a 0.4\% advantage over \textsc{Navigate}+\textsc{Answer} despite the additional tools.
This suggests that fundamental tools can yield meaningful gains, whereas realizing the benefits of a richer tool set requires adaptive orchestration that aligns tool selection and composition with intermediate observations.

\begin{wraptable}{r}{0.53\linewidth}
\vspace{-\baselineskip} % 整块表格上移一个正文行距
    \centering
    \small
    \caption{Ablation study of different training strategies.}
    \label{tab:training_ablation}
    \vspace{-0.7em}

    \setlength{\tabcolsep}{3pt}
    \renewcommand{\arraystretch}{1.05}

    \resizebox{\linewidth}{!}{
    \begin{tabular}{lcccc}
        \toprule
        \multirow{2}{*}{\textbf{Methods}}
        & \multicolumn{3}{c}{\textbf{MMLong}}
        & \multirow{2}{*}{\textbf{FinRAG}} \\
        \cmidrule(lr){2-4}
        & ANS & UNA & ALL & \\
        \midrule

        \textbf{\sys-4B}
        & \textbf{56.3} & 70.0 & \textbf{59.1}
        & \textbf{60.2} \\

        \addlinespace[3pt]
        \hspace{0.5em}-- Only Multi-ability SFT
        & 52.3 & \textbf{74.4} & 56.8 & 58.4 \\

        \bottomrule
    \end{tabular}
    }
\end{wraptable}

\noindent \textbf{Ablation on Training Strategy.}
Table~\ref{tab:training_ablation} shows that curriculum-based training outperforms direct multi-ability SFT by $2.3$ points on MMLongBench-Doc and $1.8$ points on FinRAGBench-V.
The $4.0$-point gain on answerable questions suggests that developing individual capabilities first improves their subsequent orchestration.
The accompanying $4.4$-point decline on unanswerable questions may reflect a greater tendency to answer beyond the available evidence, motivating stronger grounding verification.
These gains coincide with a shift in tool use from broad navigation toward fine-grained evidence inspection; see Appendix~\ref{sec:app_slm_tool_usage} for details.

\vspace{-.5em}
\subsection{Behavior and Robustness Analysis}
\label{sec:behavior_robustness}

\begin{figure}[h!]
    \centering

    \begin{minipage}[t]{0.54\linewidth}
        \vspace{0pt}
        \centering

        \captionof{table}{Page localization on MMLongBench-Doc.}
        \label{tab:page_localization}
        \vspace{-1em}

        \small
        \setlength{\tabcolsep}{2pt}
        \renewcommand{\arraystretch}{0.9}

        \resizebox{\linewidth}{!}{%
        \begin{tabular}{lcccc}
            \toprule
            \textbf{Method}
            & \textbf{\#Pages}$\downarrow$
            & \textbf{Rec.}$\uparrow$
            & \textbf{Prec.}$\uparrow$
            & \textbf{Acc.}$\uparrow$ \\
            \midrule

            BM25 Top-5
            & 5 & 48.5 & 14.7 & -- \\

            ColQwen2.5 Top-5
            & 5 & 74.8 & 23.5 & -- \\

            ColQwen2.5 Top-10
            & 10 & 83.9 & 14.5 & -- \\

            \midrule

            MDocAgent Retriever
            & 13.6 & 71.1 & 7.0 & 49.6 \\

            SimpleDoc Retriever
            & 4.9 & 89.0 & 34.7 & 64.0 \\

            DocLens (Gemini-2.5-Pro)
            & 3.5 & \textbf{97.3} & 55.1 & 67.6 \\

            \midrule

            \sys (GPT-5.4)
            & 4.3 & 95.0 & 55.8 & \textbf{73.5} \\

            \sys (Gemini-2.5-Pro)
            & \textbf{3.0} & 91.8 & \textbf{73.2} & 72.8 \\

            \bottomrule
        \end{tabular}%
        }

    \end{minipage}
    \hfill
    \begin{minipage}[t]{0.45\linewidth}
        \vspace{0pt}
        \centering

        \includegraphics[width=\linewidth]{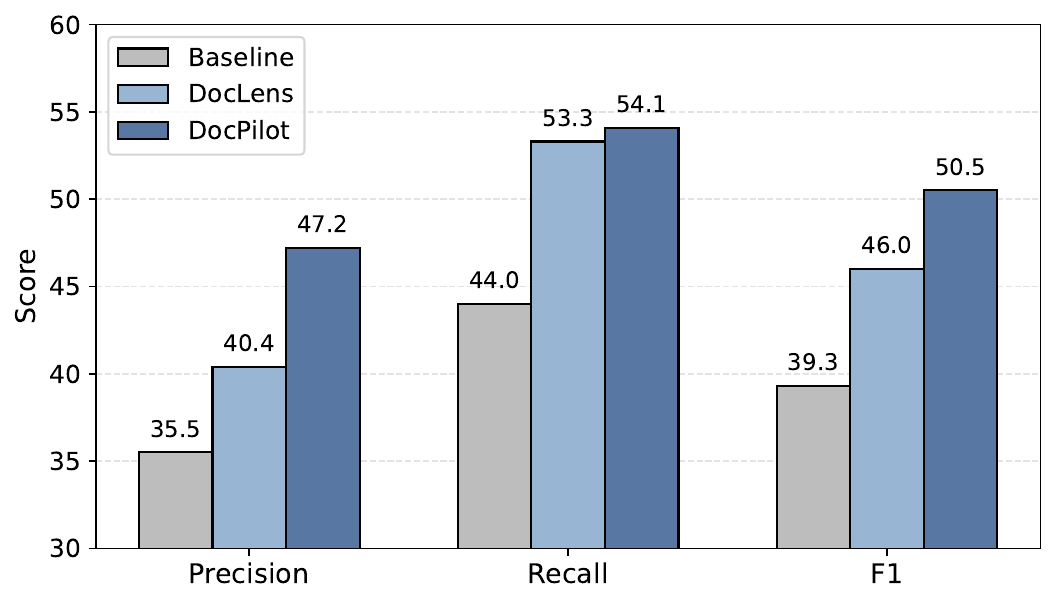}

        \vspace{-1em}

        \captionof{figure}{Element localization evaluation}
        \label{fig:element_localization}

    \end{minipage}
\vspace{-1em} 
\end{figure}

\noindent \textbf{Evidence Localization Performanc.}
% Compare DocPILOT, ColQwen top-2, MDocAgent, and SimpleDoc.
Table~\ref{tab:page_localization} reports page-level evidence localization performance. 
Compared with conventional retrievers, BM25 and ColQwen2.5~\citep{DBLP:conf/iclr/FaysseSWOVHC25}, DocLens raises page recall to 97.3 through repeated sampling with Gemini-2.5-Pro, retrieving 3.5 pages on average at 55.1 precision.
Under the same backbone, \sys retrieves only $3.0$ pages on average and raises precision to 73.2 while maintaining 91.8 recall, improving final QA accuracy from 67.6 to 72.8. With GPT-5.4, \sys also reaches 95.0 recall and 73.5 accuracy.
% See more details in Appendix~\ref{}.

Fig.~\ref{fig:element_localization} evaluates element-level localization on FinRAGBench-V using annotated evidence boxes. \sys improves precision, recall, and F1 over DocLens by +6.8, +0.8, and +4.5 points, respectively, 
showing that local inspection better matches the reasoning need than exhaustively exposing detected elements.
Together with state-conditioned page navigation, this makes evidence localization adaptive and interleaved with reasoning, with each step guiding further evidence acquisition.

\begin{figure*}[t]
    \centering

    % =========================================================
    % Left: MADQA Table
    % =========================================================
    \begin{minipage}[t]{0.63\linewidth}
        \vspace{0pt}
        \centering
        \footnotesize

        \captionof{table}{
            Performance of different systems on MADQA.
        }
        \label{tab:madqa}

        \vspace{-1em}

        \renewcommand{\arraystretch}{1.02}
        \setlength{\tabcolsep}{3.2pt}

        \resizebox{\linewidth}{!}{%
        \begin{tabular}{lccccc}
            \toprule
            \textbf{Model / Framework} 
            & \textbf{Accuracy} 
            & \textbf{X-Page} 
            & \textbf{X-Doc}
            & \textbf{Page F1} 
            & \textbf{Kuiper} $\downarrow$ \\
            \midrule
        
            \multicolumn{6}{c}{\textit{Non-Agentic Systems}} \\
            \midrule
        
            GPT-5.4 \framework{File Search}
            & 65.1\,{\scriptsize $\pm\,6.8$}
            & 36.8\,{\scriptsize $\pm\,22.6$}
            & 26.9\,{\scriptsize $\pm\,20.2$}
            & 64.5 
            & -- \\
        
            GPT-5.4 \framework{HEAVEN}
            & 71.9\,{\scriptsize $\pm\,6.4$}
            & 53.9\,{\scriptsize $\pm\,23.5$}
            & 57.4\,{\scriptsize $\pm\,24.8$}
            & 61.9 
            & -- \\
        
            GPT-5.4 \framework{Codex}
            & \cellcolor{firstBest}\textbf{86.7}\,{\scriptsize $\pm\,\textbf{5.0}$}
            & \cellcolor{firstBest}\textbf{79.4}\,{\scriptsize $\pm\,\textbf{19.6}$}
            & \cellcolor{secondBest}80.6\,{\scriptsize $\pm\,18.7$}
            & \cellcolor{firstBest}\textbf{83.0}
            & -- \\
        
            \midrule
            \multicolumn{6}{c}{\textit{Agentic Systems}} \\
            \midrule
        
            GPT-5.4 \framework{BM25 Agent}
            & 61.5\,{\scriptsize $\pm\,6.9$}
            & 48.2\,{\scriptsize $\pm\,23.5$}
            & 48.3\,{\scriptsize $\pm\,22.9$}
            & 58.3 
            & 16.7 \\
        
            GPT-5.4 \framework{RLM}
            & 60.5\,{\scriptsize $\pm\,6.9$}
            & 56.7\,{\scriptsize $\pm\,23.4$}
            & 63.8\,{\scriptsize $\pm\,24.2$}
            & 22.1 
            & 22.9 \\
        
            GPT-5.4 \framework{\sys}
            & \cellcolor{secondBest}86.5\,{\scriptsize $\pm\,5.1$}
            & \cellcolor{secondBest}59.5\,{\scriptsize $\pm\,23.2$}
            & \cellcolor{firstBest}\textbf{89.5}\,{\scriptsize $\pm\,\textbf{11.4}$}
            & \cellcolor{secondBest}75.0
            & \cellcolor{firstBest}\textbf{5.7} \\
        
            \sys-4B
            & 60.0\,{\scriptsize $\pm\,7.0$}
            & 39.7\,{\scriptsize $\pm\,23.0$}
            & 40.3\,{\scriptsize $\pm\,22.4$}
            & 50.1 
            & \cellcolor{secondBest} 10.3 \\
        
            \bottomrule
        \end{tabular}
        }

    \end{minipage}%
    \hspace{0.01\linewidth}%
    % =========================================================
    % Right: Tool-use Figure
    % =========================================================
    \begin{minipage}[t]{0.36\linewidth}
        \vspace{0pt}
        \centering

        \includegraphics[
            width=\linewidth,
            keepaspectratio
        ]{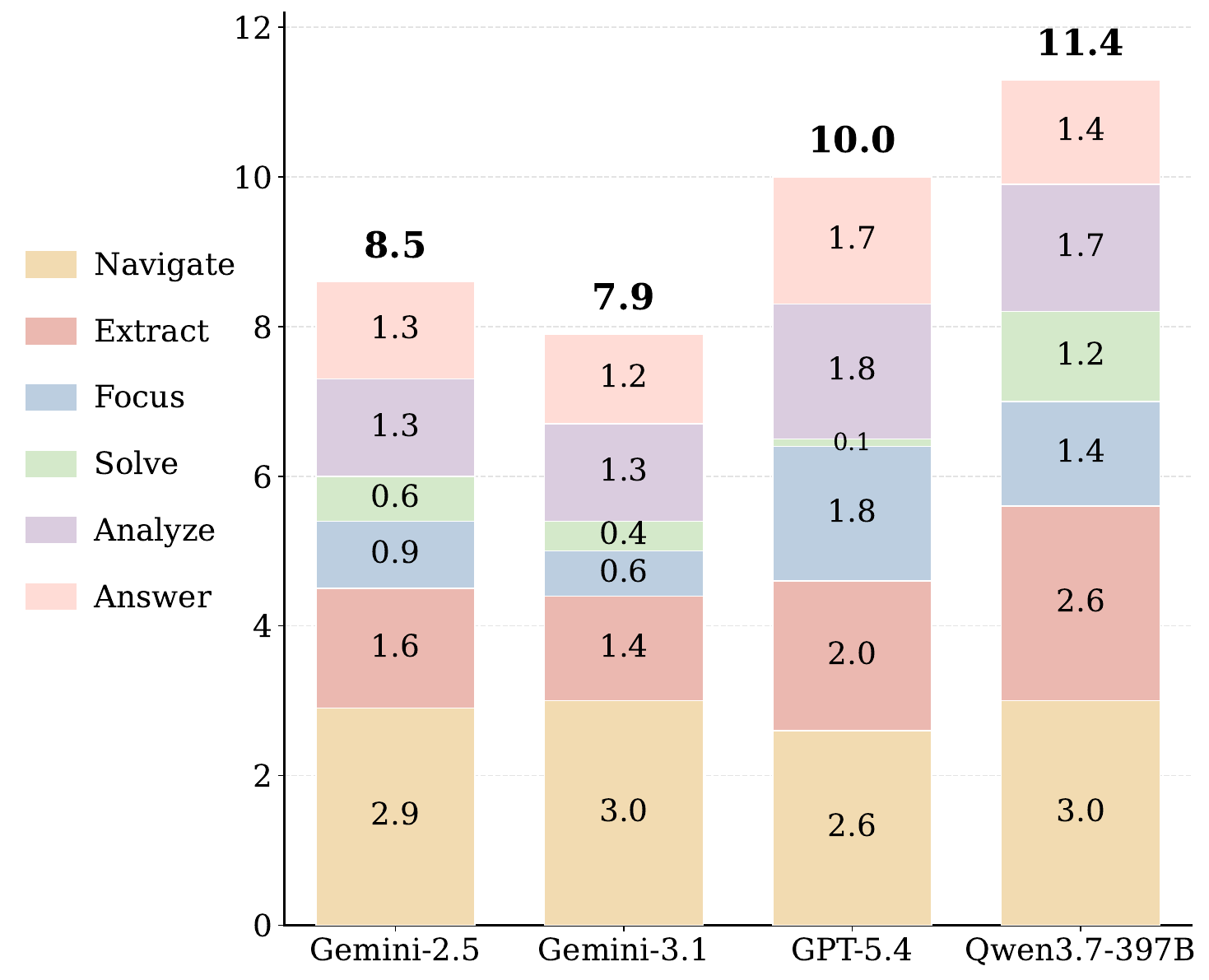}

        \vspace{-1em}

        \captionof{figure}{
            Average action calls.
        }
        \label{fig:tool_use}

    \end{minipage}

    \vspace{-2em}
\end{figure*}

\noindent \textbf{Tool-Use Analysis.}
% Analyze action frequency, action transition patterns, and tool usage across question types.
Fig.~\ref{fig:tool_use} reveals distinct orchestration patterns across backbones. \texttt{Navigate} remains consistently frequent (2.6--3.0 calls), suggesting that most behavioral variation emerges after localization. Gemini-3.1-Pro yields the shortest trajectories (7.9 actions), whereas Qwen3.7-397B-A
17B produces the longest (11.4), with heavier \texttt{<Extract>} and \texttt{<Solve>} usage. Actions are also selectively invoked: GPT-5.4 rarely invokes \texttt{<Solve>} (0.1), instead favoring \texttt{<Focus>} for fine-grained visual inspection. 
Meanwhile, repeated action calls highlight iterative evidence acquisition, inspection, and verification, while overall usage patterns reveal backbone-specific orchestration strategies.

\noindent \textbf{Cross-Document Generalization on MADQA.}
% Evaluate robustness and generalization in multi-document QA settings.
We further evaluate \sys on the 200-sample MADQA development set under an oracle-guided setting, where relevant documents are provided directly, isolating evidence localization from document retrieval.
With GPT-5.4, \sys reaches 86.5 accuracy, comparable to Codex (86.7) while surpassing the BM25 Agent (61.5) and RLM (60.5).
It performs particularly well on cross-document questions, achieving the best X-Doc score of 89.5, alongside 59.5 X-Page and 75.0 Page F1.
Moreover, \sys records the lowest Kuiper score (5.7) among agentic systems, indicating better effort calibration.
Overall, these results show strong generalization to multi-document reasoning, especially for integrating evidence across documents.

% \input{tab/madqa}

% =========================
% MADQA + Tool-use
% =========================

% \noindent \textbf{Failure Analysis.}
% Analyze typical failure cases such as wrong evidence seeking, visual misreading, computation errors, and unsupported answers.

\vspace{-.5em}
\section{Related Work}
\label{sec:related}
\vspace{-.5em}

\subsection{Multimodal Long-Document Reasoning}

Early approaches derive textual or structural representations using OCR or parsers, potentially losing critical visual cues~\citep{xu2020layoutlm}.
Vision-language models (VLMs) process page images directly but may overlook sparse evidence in long contexts and produce hallucinations~\citep{DBLP:conf/nips/MaZC0JLLLMDZP0W24}.
Retrieval-augmented methods~\citep{dong2026benchmarking,DBLP:conf/mm/ZhangYZ24} scale by retrieving relevant text~\citep{DBLP:journals/corr/abs-2402-03216,DBLP:conf/naacl/SanthanamKSPZ22} or visual pages~\citep{DBLP:conf/iclr/FaysseSWOVHC25} before reasoning (e.g., ViDoRAG~\citep{DBLP:conf/emnlp/WangDCWWXZ25}, M3DocRAG~\citep{DBLP:journals/corr/abs-2411-04952}, and VRAG-RL~\citep{DBLP:conf/nips/WangDZCCWXHZ25}).
Yet their static, similarity-driven retrieval may miss evidence required for reasoning.

\vspace{-.3em}
Agentic frameworks enable multi-step document interaction.
MDocAgent~\citep{DBLP:journals/corr/abs-2503-13964} routes evidence through specialized agents; SimpleDoc~\citep{DBLP:conf/emnlp/JainWZLDSWW25} alternates dual-cue retrieval with query refinement; and DocLens~\citep{DBLP:conf/acl/ZhuMCLPY26} follows a coarse-to-fine pipeline of repeated page localization and exhaustive element enrichment.
More flexible approaches include DocDancer~\citep{DBLP:journals/corr/abs-2601-05163}, which iteratively seeks information through \textsc{Search} and \textsc{Read}, and Doc-V*~\citep{DBLP:conf/acl/ZhengFLWZRZWLLC26}, which aggregates page-level evidence without OCR.
However, prescribed workflows or limited action spaces often constrain adaptive orchestration.
\sys instead coordinates fine-grained actions for evidence seeking, multimodal understanding, and iterative refinement.

\vspace{-.5em}
\subsection{Trainable Document Agents}
\vspace{-.5em}

% Recent work has explored training document agents to learn interaction policies from trajectories and environment feedback~\citep{DBLP:journals/corr/abs-2510-16872,DBLP:journals/corr/abs-2604-12812,DBLP:conf/nips/WangDZCCWXHZ25,DBLP:journals/corr/abs-2503-09516}. 
% Given the scarcity of agentic supervision in DocQA, synthetic interaction trajectories provide a scalable alternative to costly manual annotation~\citep{liu2025comprehensive,Qwen37Plus}.
% DocDancer~\citep{DBLP:journals/corr/abs-2601-05163} learns \textsc{Search}/\textsc{Read} behaviors from synthesized trajectories, while Doc-V*~\citep{DBLP:conf/acl/ZhengFLWZRZWLLC26} combines trajectory imitation with reinforcement learning for page-level navigation. 
% However, trajectory-level supervision induces a fundamental learning mismatch: coarse workflows simplify policy learning while leaving action-level multimodal capabilities weakly supervised, whereas more expressive action spaces expose these capabilities but make their composition harder.
% \sys addresses this through data-grounded trajectory synthesis and curriculum-based agentic training, progressing from single-ability fine-tuning to multi-ability learning.

Recent studies train document agents' interaction policies from trajectories and environment feedback~\citep{DBLP:journals/corr/abs-2510-16872,DBLP:journals/corr/abs-2604-12812,DBLP:conf/nips/WangDZCCWXHZ25,DBLP:journals/corr/abs-2503-09516}.
With limited agentic supervision in DocQA, synthetic trajectories offer a scalable alternative to manual annotation~\citep{liu2025comprehensive,Qwen37Plus}.
DocDancer~\citep{DBLP:journals/corr/abs-2601-05163} learns \textsc{Search}/\textsc{Read} behaviors from synthesized trajectories, while Doc-V*~\citep{DBLP:conf/acl/ZhengFLWZRZWLLC26} combines trajectory imitation with reinforcement learning for page navigation.
However, coarse-grained trajectories often weakly supervise multimodal capabilities, whereas fine-grained actions complicate learning adaptive composition.
\sys addresses this tension through data-grounded trajectory synthesis and curriculum-based agentic training, progressing from single-ability acquisition to multi-ability orchestration.
\vspace{-.5em}
\section{Conclusion and Future Work}
\label{sec:conclusion}
\vspace{-.5em}

In this work, we presented \sys, an agentic document framework that unifies adaptive evidence seeking, multimodal understanding, and iterative refinement within a shared document environment. Data-grounded trajectory synthesis supports a curriculum from single-ability fine-tuning to multi-ability agentic training, enabling compact models to learn both individual capabilities and their adaptive composition. Evaluations on four benchmarks show substantial gains across strong VLM backbones and improved evidence localization. 
Ablations further highlight the importance of orchestrating complementary tools to coordinate evidence seeking and reasoning.

% \noindent \textbf{Limitations.} 

% Looking ahead, \sys offers a foundation for more autonomous and adaptive document intelligence. Future work will extend its scope from reasoning over preselected documents to autonomous retrieval and reasoning across entire document collections. We will also explore agent self-evolution by using accumulated interaction trajectories to distill successful strategies and lessons from failures, while building an evolving semantic layer that consolidates domain knowledge across tasks.
Looking ahead, \sys provides a foundation for autonomous, adaptive document intelligence.
% We will extend it from preselected documents to autonomous retrieval and reasoning across entire document collections.
We will explore self-evolution by distilling successful strategies and failure lessons from accumulated trajectories, alongside an evolving semantic layer that consolidates domain knowledge across tasks.

\section*{AI Use Statement}
\vspace{-0.5em}
LLMs are used only for auxiliary purposes such as language refinement, minor code debugging, synthetic data construction, and experiment evaluation support. They do not contribute to the research design, methodology, or core writing of the paper. Accordingly, LLM usage does not constitute a substantive contribution to the intellectual content of this work.

\section*{Ethics Statement}
\vspace{-0.5em}
Our experiments use publicly available document QA datasets. \sys advances evidence-grounded document understanding, while its compact model supports resource-efficient, on-premise deployment. Generated answers may contain errors and require verification against source evidence, particularly in high-stakes domains such as finance and law. Deployments involving sensitive documents should incorporate appropriate access controls and privacy safeguards.
\vspace{-0.5em}

\section*{Reproducibility Statement}
\vspace{-0.5em}
To support reproducibility, we provide the implementation of \sys, including document preprocessing, agent orchestration, trajectory synthesis, and evaluation scripts, at \url{https://github.com/HKUSTDial/DocPilot}. The appendix details data construction procedures, and evaluation protocols for reproducing our experiments.
\vspace{-0.5em}


\bibliography{iclr2027_conference}
\bibliographystyle{iclr2027_conference}

\appendix
\section{Dataset Statistics}
\label{app:dataset_statistics}
% 各个评估bm的统计

We evaluate \sys on four benchmarks spanning three document reasoning settings: general-purpose understanding, domain-specific financial analysis, and cross-document reasoning. Table~\ref{tab:dataset_statistics} summarizes their key statistics.

\begin{table*}[t]
    \centering
    \small
    \caption{Statistics and characteristics of the evaluation benchmarks.}
    \label{tab:dataset_statistics}
    \renewcommand{\arraystretch}{1.05}
    \setlength{\tabcolsep}{5pt}

    \begin{tabular}{lcc|cc|cc|cc}
        \toprule

        \multirow{2}{*}{\textbf{Benchmarks}}
        & \multicolumn{2}{c|}{\textbf{Document}}
        & \multicolumn{2}{c|}{\textbf{Question}}
        & \multicolumn{2}{c|}{\textbf{Evi. Loc.}}
        & \multicolumn{2}{c}{\textbf{Doc. Input}} \\

        & Domain
        & \#Pages
        & \#Num
        & Expert
        & Page
        & Quote
        & Single
        & Multi \\

        \midrule

        MMLongBench-Doc
        & Multiple
        & 47.5
        & 1,082
        & \cmark
        & \cmark
        & \xmark
        & \cmark
        & \xmark \\

        DocBench
        & Multiple
        & 66.0
        & 1,102
        & \cmark
        & \xmark
        & \xmark
        & \cmark
        & \xmark \\

        FinRAGBench-V
        & Financial
        & 50.7
        & 1,394
        & \cmark
        & \cmark
        & \xmark
        & \cmark
        & \xmark \\

        MADQA
        & Multiple
        & 23.3
        & 2,250
        & \cmark
        & \cmark
        & \xmark
        & \cmark
        & \cmark \\

        \bottomrule
    \end{tabular}
\end{table*}

\noindent\textbf{General-Purpose Understanding.}
We adopt MMLongBench-Doc~\citep{DBLP:conf/nips/MaZC0JLLLMDZP0W24} and
DocBench~\citep{zou2025docbench} as two challenging general-purpose benchmarks.
MMLongBench-Doc contains 135 long PDFs across seven document types, averaging
47.5 pages and 21,214 textual tokens. Its refined evaluation set includes
1,082 questions, with 372 cross-page and 223 unanswerable cases.
The benchmark spans text, tables, charts, figures, and layout structures,
providing a comprehensive testbed for long-context evidence seeking and multimodal reasoning.

DocBench contains 229 real-world PDFs and 1,102 questions across five domains:
Academia, Finance, Government, Law, and News. The documents average 66 pages
and 46,377 tokens, while the questions span text-only, multimodal, metadata,
and unanswerable settings. It further exhibits substantial document-length
variation, from single-page news articles to financial reports averaging 192
pages. Together, MMLongBench-Doc and DocBench provide complementary coverage
of general long-document understanding across diverse layouts, modalities, and
reasoning requirements.

\noindent\textbf{Domain-Specific Financial Analysis.}
We further evaluate on FinRAGBench-V~\citep{zhao2025finragbench}, a multimodal benchmark built from real-world financial documents. 
Its retrieval corpus contains 2,209 documents and 111,999 pages spanning financial reports, statements, prospectuses, academic papers, magazines, and news. 
We evaluate on the 539 English QA pairs, covering text inference, chart understanding, numerical calculation, table reasoning, comparison, and multi-page queries. Its dense financial tables, charts, and fine-grained visual evidence make it well suited for evaluating precise evidence localization and domain-specific multimodal reasoning.

\noindent\textbf{Cross-Document Reasoning.}
Finally, we use the 200-question development set of MADQA~\citep{borchmann2026strategic} to evaluate reasoning over document collections. 
Unlike single-document benchmarks, MADQA requires agents to navigate document collections and aggregate dispersed evidence, making it well suited for evaluating cross-document evidence seeking and reasoning.

\section{Construction of \sys}
\label{app:construction}
% Detail of Environment Design
% Agent–Environment Interaction  Protocol
% prompt of different modules

\subsection{Detail of Environment Design}

This subsection provides additional construction details for the multimodal document environment introduced in Sec.~\ref{ssec:environment}, including the OCR-derived structure, page thumbnails, and document overview.

\noindent\textbf{OCR-derived Structured Representation.}
To support information extraction~\citep{liang2026beyond,zhang2026document},
we employ MinerU2.5~\citep{niu2026mineru2} for OCR and layout parsing, representing each page as structured elements rather than flattened text.
For page $I_i$, the representation
$\mathcal{S}_i=\{s_{i,1},\ldots,s_{i,M_i}\}$
contains elements such as titles, paragraphs, tables, figures, captions, and footnotes.
Each element retains its recognized content, semantic type, reading order, bounding box, and page index, preserving both the semantic content and spatial structure of the original page.
Structurally irrelevant elements, such as headers and footers, are removed during parsing.

\noindent\textbf{Page Thumbnail Representation.}
For each page $I_i$, the environment maintains both a high-resolution page screenshot and a lightweight thumbnail $T_i$.
The page screenshot is capped at $1024\times768$ pixels, while $T_i$ is resized to $512\times512$ and annotated with its absolute page number.
Although detailed text may be difficult to read in the thumbnail, it preserves coarse visual cues such as headings, layout, and the distribution of tables, charts, and figures, while the original screenshot supports subsequent fine-grained extraction and inspection.

\noindent\textbf{Document Overview Representation.}
To provide a compact document-level representation, we further construct a document overview $\widetilde{\mathcal{D}}$ from the page thumbnails.
Consecutive thumbnails are arranged in reading order into a regular grid, with their page indices retained for unambiguous correspondence with the original document.
For documents that exceed the capacity of a single grid, the thumbnails are partitioned into consecutive groups and represented by multiple overview images.
For multi-document inputs, an overview is constructed independently for each document and annotated with its document identifier and title when available.
The resulting representation preserves the global organization of the document, including section transitions and the distribution of visually salient content, at a substantially lower visual cost than full-resolution page images.

\subsection{Agent--Environment Interaction Protocol}
\label{app:interaction}

% thought-action-observation loop图.

We adopt the vanilla ReAct framework~\citep{yao2022react}, which interleaves reasoning and acting within a unified interaction loop.
At each step, the agent observes the question, environment state, interaction history, and available actions, and selects the next action.
The environment executes the action, returns the corresponding observation, and updates its state for the subsequent decision.

\paragraph{Action Space.}
\sys exposes six actions: \texttt{<Analyze>}, \texttt{<Navigate>}, \texttt{<Extract>}, \texttt{<Focus>}, \texttt{<Solve>}, and \texttt{<Answer>}.
\texttt{<Analyze>} decomposes the question, updates the solving plan, assesses evidence sufficiency, diagnoses failures, and verifies candidate answers.
\texttt{<Navigate>} inspects page thumbnails or document overviews to identify potentially relevant pages.
\texttt{<Extract>} selects task-relevant evidence from OCR-derived text, tables, visual elements, and page images.
\texttt{<Focus>} performs fine-grained visual inspection through cropping, zooming, or subfigure isolation.
\texttt{<Solve>} generates executable Python code over the collected evidence, which the environment executes to return the results.
\texttt{<Answer>} synthesizes a candidate response from the accumulated evidence.

\paragraph{State-conditioned Action Availability.}
The available action set is dynamically determined by the current environment state, which records navigation progress, candidate and processed pages, and candidate answers with their verification outcomes. 
These states jointly determine which actions are valid at each step. 
For example, \texttt{<Extract>} requires selected pages, \texttt{<Focus>} requires available visual evidence, and \texttt{<Solve>} requires extracted evidence for computation. 
When the next step is ambiguous or evidence is insufficient, the agent may invoke \texttt{<Analyze>} to formulate or revise its plan; otherwise, it can directly execute the appropriate action without redundant analysis.
This state-conditioned interface prevents invalid operations while allowing flexible composition of different capabilities.

\paragraph{Execution and State Update.}

Before execution, the environment validates the selected action, its arguments, and referenced page or evidence identifiers.
Invalid actions or references leave the environment unchanged.
After successful execution, the resulting observation is appended to the interaction history, while relevant states, including candidate pages, extracted evidence, focused regions, computation results, candidate answers, and verification feedback, are updated accordingly.
This persistent state allows subsequent decisions to build directly on previous interactions.

\paragraph{Verification and Termination.}
\texttt{<Answer>} produces a candidate response, which can be verified by \texttt{<Analyze>} against the question and accumulated evidence.
If accepted, the interaction terminates; otherwise, identified evidence gaps or inconsistencies guide further evidence seeking, reasoning, computation, or answer revision.
The interaction also stops at the maximum step budget, returning ``INSUFFICIENT\_EVIDENCE`` if no sufficiently supported answer is obtained.

\subsection{Module Design Details}

This appendix exposes four specialized tools with explicit descriptions and structured argument schemas.
\texttt{<Navigate>} scans page thumbnails to locate potentially relevant pages, while \texttt{<Extract>} retrieves question-relevant evidence from the selected pages.
When finer visual interpretation is required, \texttt{<Focus>} applies targeted operations such as zooming, cropping, and subfigure isolation.
For numerical or symbolic reasoning, \texttt{<Solve>} generates and executes Python code over the collected evidence.
At each step, the agent conditions on the tool descriptions, current evidence state, and interaction history to determine which tool to invoke and how to instantiate its arguments, enabling adaptive tool selection without a fixed execution sequence.
Detailed prompts and implementation specifications for each tool are provided in Appendix~\ref{sec:app_tool_prompt}.

\definecolor{figpurple}{HTML}{A18AB8}

\begin{tcolorbox}[
    colback=figpurple!5,
    colframe=figpurple,
    colbacktitle=figpurple,
    coltitle=white,
    title=\textbf{Tool schema},
    fonttitle=\small\bfseries,
    fontupper=\small,
    arc=2pt,
    outer arc=2pt,
    left=2mm,
    right=2mm,
    top=2mm,
    bottom=2mm,
    enhanced,
    breakable
]

\textbf{\texttt{<Navigate>}.} Inspect page thumbnails to identify relevant pages and return candidates with reasons.

\vspace{0.2em}
\hrule
\vspace{0.2em}

\begin{tabular}{p{0.18\linewidth} p{0.18\linewidth} p{0.56\linewidth}}
\textbf{Field} & \textbf{Type} & \textbf{Description} \\
\hline
\texttt{query} & string & Navigation goal or analyzed question. \\
\texttt{doc\_id} & string & Document ID. \\
\texttt{pages} & string & Optional pages or ranges to inspect. \\
\texttt{force\_refresh} & boolean & Reinspect previously verified pages if enabled. \\
\texttt{refresh\_reason} & string & Reason for refreshing previously verified pages. \\
\end{tabular}

\vspace{0.2em}
\hrule
\vspace{0.8em}

\textbf{\texttt{<Extract>}.} Extract question-relevant evidence from selected pages.

\vspace{0.2em}
\hrule
\vspace{0.2em}

\begin{tabular}{p{0.18\linewidth} p{0.18\linewidth} p{0.56\linewidth}}
\textbf{Field} & \textbf{Type} & \textbf{Description} \\
\hline
\texttt{pages} & string & Candidate pages returned by navigation \\
\texttt{instruction} & string & Task-specific extraction goal. \\
\texttt{max\_pages} & integer & Maximum number of pages to process. \\
\end{tabular}

\vspace{0.2em}
\hrule
\vspace{0.8em}

\textbf{\texttt{<Focus>}.} Perform fine-grained visual inspection on extracted evidence.

\vspace{0.2em}
\hrule
\vspace{0.2em}

\begin{tabular}{p{0.18\linewidth} p{0.18\linewidth} p{0.56\linewidth}}
\textbf{Field} & \textbf{Type} & \textbf{Description} \\
\hline
\texttt{instruction} & string & Visual inspection goal. \\
\texttt{figure\_refs} & array & Figure references from extraction. \\
\texttt{operations} & array & Visual operations for the target figure, e.g., zoom, crop, or subfigure isolation. \\
\end{tabular}

\vspace{0.2em}
\hrule
\vspace{0.8em}

\textbf{\texttt{<Solve>}.} Generate Python code for deterministic reasoning over collected evidence.

\vspace{0.2em}
\hrule
\vspace{0.2em}

\begin{tabular}{p{0.18\linewidth} p{0.18\linewidth} p{0.56\linewidth}}
\textbf{Field} & \textbf{Type} & \textbf{Description} \\
\hline
\texttt{instruction} & string & Code-generation or computation goal. \\
\texttt{evidence\_ids} & array & Evidence items used for computation. \\
\end{tabular}

\vspace{0.2em}
\hrule

\end{tcolorbox}

\section{Trajectory Synthesis Details}
\label{sec:app_synthesis}
% 讲一下怎么合成
% case表示小模型生成的轨迹质量

\begin{table}[t]
    \centering
    \caption{Datasets used for data-grounded trajectory synthesis and their capability coverage.}
    \label{tab:trajectory_sources}
    \resizebox{\linewidth}{!}{
    \begin{tabular}{l|l|c|l}
        \toprule
        \multicolumn{2}{c|}{\textbf{Type}}
        & \textbf{Source Datasets}
        & \textbf{Primary Capabilities} \\
        \midrule
        \multirow{4}{*}{\shortstack[c]{Single-\\ability}}
        & \texttt{<Navigate>}
        & DocSeeker~\citep{yan2026docseeker}
        & Page-level navigation \\
        
        & \texttt{<Extract>}
        & HotpotQA~\citep{yang2018hotpotqa}
        & Evidence localization \\

        & \texttt{<Solve>}
        & File-CSV~\citep{DBLP:journals/corr/abs-2510-16872}
        & Code Generation \\
        
        & \texttt{<Answer>}
        & FinQA~\citep{chen2021finqa}
        & Table reasoning \\
        
        & \texttt{<Answer>}
        & ChartQAPro~\citep{masry2025chartqapro}, InfographicVQA~\citep{mathew2022infographicvqa}
        & Chart and visual reasoning \\
        
        \midrule
        \multicolumn{2}{c|}{\multirow{2}{*}{Multi-ability}}
        & \multicolumn{1}{c|}{LongDocURL~\citep{deng2025longdocurl}, DUDE~\citep{van2023document}}
        & \multicolumn{1}{l}{Long-document, multimodal, and} \\
        \multicolumn{2}{c|}{}
        & \multicolumn{1}{c|}{MADQA~\citep{borchmann2026strategic}, MMDocRAG~\citep{dong2026benchmarking}}
        & \multicolumn{1}{l}{cross-document reasoning} \\
        \bottomrule
    \end{tabular}
    }
\end{table}

% \begin{table}[t]
%     \centering
%     \caption{Statistics of \textsc{DocAgent-Instruct-40K}.}
%     \label{tab:trajectory_statistics}
%     \begin{tabular}{llr}
%         \toprule
%         \textbf{Trace Type}
%         & \textbf{Capability}
%         & \textbf{Number} \\
%         \midrule
%         \multirow{4}{*}{Single-ability}
%         & \texttt{<Navigate>} & 13,986 \\
%         & \texttt{<Answer>}   & 10,749 \\
%         & \texttt{<Extract>}  & 6,196 \\
%         & \texttt{<Solve>}    & 3,007 \\
%         \cmidrule(lr){2-3}
%         & \textit{Subtotal}   & 33,938 \\
%         \midrule
%         Multi-ability
%         & Interaction traces  & 4,752 \\
%         \midrule
%         \textbf{Total}
%         &                     & \textbf{38,690} \\
%         \bottomrule
%     \end{tabular}
% \end{table}

\begin{figure*}[t]
    \centering
    \begin{minipage}[c]{0.52\textwidth}
        \centering
        \small
        \captionof{table}{Statistics of \textsc{DocAgent-Instruct-40K}.}
        \label{tab:trajectory_statistics}
        \vspace{-0.8em}
        \begin{tabular}{llr}
            \toprule
            \textbf{Trace Type} & \textbf{Capability} & \textbf{Number} \\
            \midrule
            \multirow{4}{*}{Single-ability}
            & \texttt{<Navigate>} & 13,986 \\
            & \texttt{<Answer>}   & 10,749 \\
            & \texttt{<Extract>}  & 6,196 \\
            & \texttt{<Solve>}    & 3,007 \\
            \cmidrule(lr){2-3}
            & \textit{Subtotal}  & 33,938 \\
            \midrule
            Multi-ability & Interaction traces & 4,752 \\
            \midrule
            \textbf{Total} & & \textbf{38,690} \\
            \bottomrule
        \end{tabular}
    \end{minipage}\hfill
    \begin{minipage}[c]{0.46\textwidth}
        \centering
        \includegraphics[width=\linewidth]{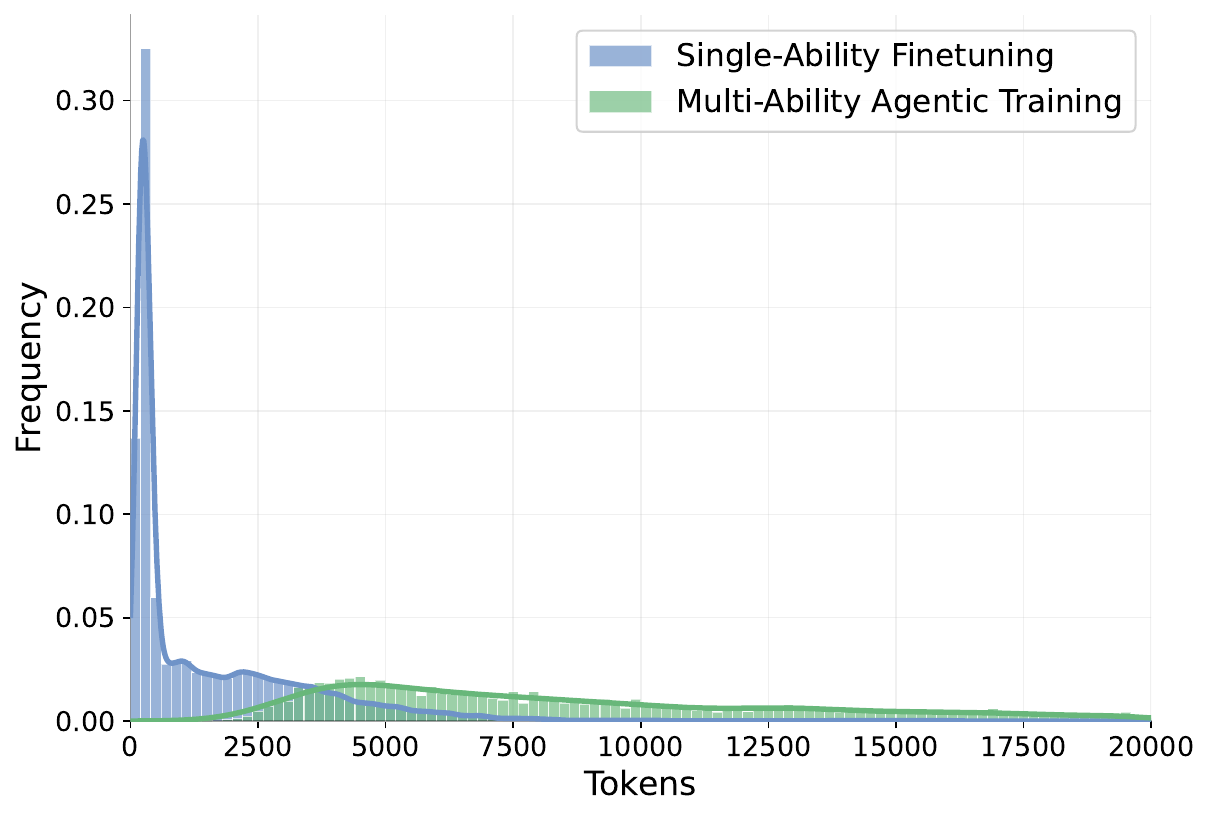}
        \par\vspace{-1em}
        \captionof{figure}{Length distribution of training data.}
        \label{fig:token_distribution}
    \end{minipage}
    \vspace{-1em}
\end{figure*}

% \begin{figure}[h!]
%     \centering
%     \includegraphics[width=0.6\linewidth]{fig/token_distribution.pdf}
%     \vspace{-1em}
%     \caption{
%         Length distribution of training data.
%     }
%     \label{fig:navigation_process}
%     \vspace{-1em}
%  \end{figure}

\noindent\textbf{Motivation.}
Existing DocVQA datasets predominantly provide document--question--answer triplets, without supervising the intermediate decisions required for interactive document solving.
This supervision is insufficient for learning both the execution of individual tools and their adaptive composition over long trajectories.
Collecting such traces manually is also costly, as it requires repeated document inspection, evidence annotation, tool execution, and answer verification.
We therefore construct \textsc{DocAgent-Instruct-40K} through data-grounded trajectory synthesis, using existing annotations to constrain teacher generation and ensure that the resulting traces remain executable and evidence-grounded.

\noindent\textbf{Data Sources.}
As shown in Table~\ref{tab:trajectory_sources}, we select datasets according to the capabilities required by \sys.
For single-ability supervision, DocSeeker provides page-level evidence annotations for \texttt{<Navigate>}, while HotpotQA supplies multi-hop evidence-seeking examples for \texttt{<Extract>}.
For \texttt{<Solve>}, we use the \texttt{file\_csv} subset of DeepAnalyze, where each instance pairs a table and question with target code, providing supervision for code-based reasoning.
For \texttt{<Answer>}, we synthesize reasoning traces from QA datasets spanning multiple modalities: FinQA for table reasoning, ChartQAPro for chart reasoning, and InfographicVQA for visually rich document understanding.
These datasets are used to synthesize focused reasoning traces for individual capabilities.

For multi-ability supervision, we use existing DocVQA benchmarks that require longer and more heterogeneous interaction.
LongDocURL contains long-document questions with dispersed evidence; DUDE covers diverse document layouts and question types; MADQA emphasizes cross-page and cross-document reasoning; and MMDocRAG requires retrieval and reasoning over multimodal document collections.
Together, these datasets provide realistic tasks for synthesizing complete interaction trajectories involving evidence seeking, multimodal reasoning, and verification.

\noindent\textbf{Teacher-Guided Synthesis.}
We use Qwen3.5-397B-A17B as our multimodal synthesis teacher.

For \textit{reasoning trace synthesis}, we construct capability-specific prompts from the question, task input, and target output.
The teacher generates focused traces for locating evidence-bearing pages, extracting grounded evidence, answering questions across document modalities, or producing executable Python code.
Generated code is executed to validate its result, while other traces are checked against the reference answers or evidence annotations.

For \textit{interaction trace synthesis}, the teacher is rolled out within the full \sys environment.
At step \(t\), it selects an action and its arguments conditioned on the question \(q\), interaction history \(H_t\), and accumulated evidence \(E_t\):
\begin{equation}
    (a_t,\phi_t)
    \sim
    \mathcal{T}\!\left(q,H_t,E_t\right),
\end{equation}
where \(\mathcal{T}\) denotes the teacher model.
The environment executes the action and returns observation \(o_t\), after which the evidence state and interaction history are updated:
\begin{equation}
   E_{t+1}=\operatorname{Update}(E_t,o_t),
    \qquad
    H_{t+1}=H_t\cup\{(a_t,\phi_t,o_t)\}.  
\end{equation}
This process continues until the teacher produces a verified answer or reaches the interaction budget.
No fixed action sequence is imposed: the teacher may revisit navigation, extraction, visual inspection, or computation as new evidence gaps emerge.

We apply three quality-control criteria to the resulting trajectories.
First, \textit{answer correctness} requires the final prediction to match the reference answer under the corresponding dataset metric.
Second, \textit{interaction validity} checks tool names, argument schemas, document references, and executable code.
Third, \textit{evidence grounding} verifies that the answer is supported by the evidence collected during interaction.
We further discard malformed trajectories, repeated action loops, and executions containing no useful evidence acquisition.

\noindent\textbf{Data Format and Statistics.}
Table~\ref{tab:trajectory_statistics} summarizes the dataset composition, while Fig.~\ref{fig:token_distribution} presents the token-length distributions across the two training stages. After filtering, \textsc{DocAgent-Instruct-40K} contains approximately 40K examples: 33,938 single-ability reasoning traces for developing individual document-solving capabilities and 4,752 multi-ability interaction traces for learning their adaptive composition over complete trajectories. We set the maximum sequence lengths to 8K and 20K text tokens for the two stages, respectively, excluding visual tokens.

% \section{Detail of Curriculum Training}
% 还要讲一下inference

\section{More Evaluation Results}

% evidence累计过程
% tool use
% 不同方法对于文档长度的效果图

\subsection{Evidence Recall across Navigation Rounds}
\label{app:cumulative_recall}

We examine how \sys accumulates page-level evidence over successive navigation rounds on MMLongBench-Doc~\citep{DBLP:conf/nips/MaZC0JLLLMDZP0W24}.
Each \texttt{<Navigate>} action constitutes one round.
For question $i$, let $P_{i,t}$ denote the pages retrieved at round $t$ and $G_i$ its ground-truth evidence pages.
We measure macro-average cumulative recall as
\begin{equation}
R_t =
\frac{1}{|\mathcal{I}|}
\sum_{i\in\mathcal{I}}
\frac{\left|\left(\bigcup_{s=1}^{t}P_{i,s}\right)\cap G_i\right|}
{|G_i|},
\end{equation}
where $\mathcal{I}$ contains questions with nonempty evidence annotations.

\begin{wrapfigure}{r}{0.51\linewidth}
    \vspace{-\intextsep} % 抵消顶部默认间距
    \centering
    \includegraphics[width=\linewidth]{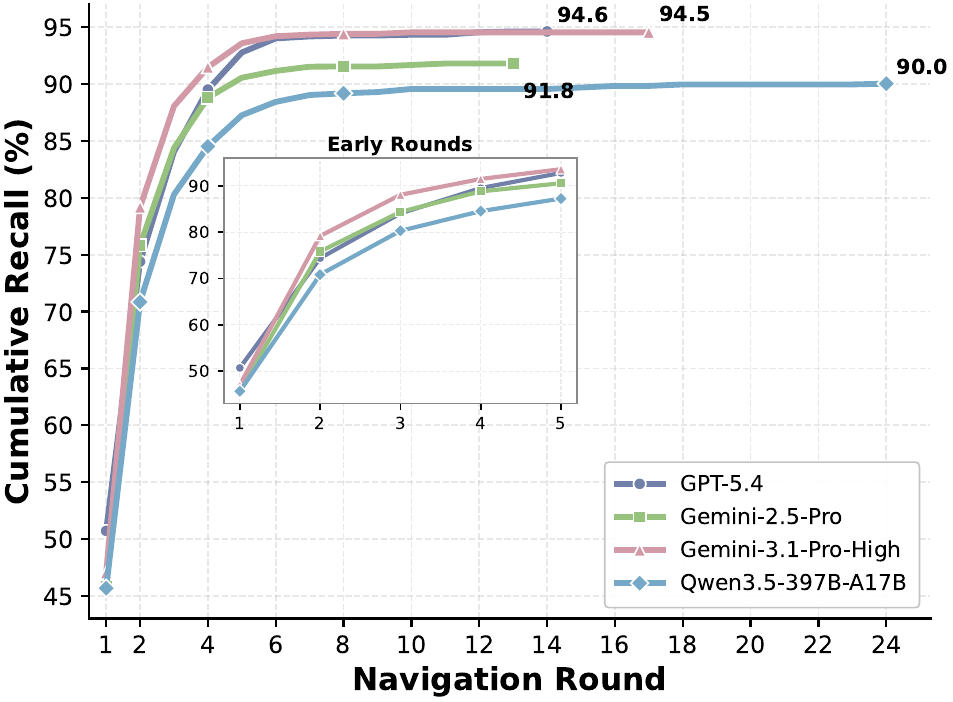}
    \vspace{-2em}
    \caption{Curriculum-based training.}
    \label{fig:curriculum_training}
    \vspace{-1em}
\end{wrapfigure}
\noindent\textbf{Progressive evidence acquisition.}
Across backbones, first-round recall ranges from $45.67\%$ to $50.71\%$, leaving substantial evidence undiscovered.
By the third round, recall reaches $80.29\%$--$88.06\%$, gaining $33.34$--$41.04$ percentage points over the first round.
Final recall rises to $90.04\%$--$94.63\%$.
These gains show that evidence acquisition extends substantially beyond the initial navigation decision, supporting adaptive interaction for broader evidence coverage.

\noindent\textbf{Diminishing returns.}
Evidence coverage expands rapidly in the early rounds, with all backbones reaching $96.9\%$--$99.0\%$ of their final recall by round five.
Navigation beyond round six adds only $0.34$--$1.60$ percentage points, indicating diminishing returns at greater navigation depths.
Despite this aggregate saturation, individual questions may still require further evidence seeking.

\noindent\textbf{Backbone-dependent convergence.}
Gemini-3.1-Pro achieves higher recall than GPT-5.4 from rounds two through six, while both converge to similar final coverage ($94.55\%$ versus $94.63\%$).
Qwen3.5-397B-A17B improves more gradually, reaching $90.04\%$ by round 24.
Thus, navigation depth alone does not determine evidence coverage: backbones differ in both how quickly they acquire relevant evidence and the coverage they ultimately achieve.

\subsection{Tool Usage Analysis of \sys-4B}
\label{sec:app_slm_tool_usage}

\begin{table}[t]
    \centering
    \caption{Average number of tool calls under different training strategies on MMLongBench-Doc.}
    \vspace{-1em}
    \label{tab:tool_usage}
    \resizebox{\linewidth}{!}{
    \begin{tabular}{lccccccc}
        \toprule
        \textbf{Training Methods}
        & \textbf{Navigate}
        & \textbf{Extract}
        & \textbf{Focus}
        & \textbf{Solve}
        & \textbf{Analyze}
        & \textbf{Answer}
        & \textbf{Total} \\
        \midrule
        \textbf{Curriculum-based Agentic Training}
              & 2.87 & 2.07 & 1.46 & 0.85 & 8.20 & 0.95 & 16.41 \\
          \quad -Only Multi-ability Fine-tuning
              & 3.75 & 2.03 & 0.74 & 0.71 & 8.21 & 0.98 & 16.43 \\
        \bottomrule
    \end{tabular}
    }
\end{table}

Table~\ref{tab:tool_usage} compares the tool usage of curriculum-based agentic training with direct multi-ability fine-tuning.
The two strategies yield nearly identical total call counts (16.41 vs.\ 16.43), while \texttt{<Extract>}, \texttt{<Analyze>}, and \texttt{<Answer>} usage remains largely unchanged.
The main difference therefore lies in how interactions are allocated across specialized capabilities.

Curriculum training reduces \texttt{<Navigate>} calls from 3.75 to 2.87, nearly doubles \texttt{<Focus>} calls from 0.74 to 1.46, and increases \texttt{<Solve>} calls from 0.71 to 0.85.
This redistribution indicates a shift from broad page exploration toward local visual inspection and code-based reasoning over collected evidence.
At a comparable number of interactions, the curriculum-trained agent devotes more steps to examining and reasoning over candidate evidence.

These patterns are consistent with the curriculum's progression from individual capability acquisition to adaptive orchestration.
Single-ability fine-tuning provides a foundation for subsequent tool composition, potentially helping the agent move beyond repeated navigation toward finer-grained evidence processing.
This offers a plausible explanation for the curriculum's performance gains, although call frequencies alone do not establish whether each invocation is appropriate or successful.

 \begin{table*}[t]
    \centering
    \footnotesize
    \caption{Performance comparison on MMLongBench-Doc by document type. \textbf{Bold} indicates the best result, and \underline{underlined} indicates the second best. Parenthesized values for \sys-4B denote percentage-point changes over vanilla Qwen3.5-4B; green indicates gains and red indicates declines.}
    \vspace{-1em}
    \label{tab:domain_results}
    \setlength{\tabcolsep}{4.0pt}
    \renewcommand{\arraystretch}{1.05}

    \resizebox{\linewidth}{!}{
    \begin{tabular}{l ccccccc|c}
    \toprule
    \multirow{2}{*}{\textbf{Model}}
    & \multicolumn{7}{c|}{\textbf{Document Type \& Domain}}
    & \multirow{2}{*}{\textbf{Overall}} \\
    \cmidrule(lr){2-8}
    & \textbf{Research}
    & \textbf{Acad. Paper}
    & \textbf{Guidebook}
    & \textbf{Tutorial}
    & \textbf{Finance}
    & \textbf{Brochure}
    & \textbf{Industry}
    & \\
    \midrule

    Qwen3.5-4B
  & 54.6 & 52.0 & 54.6 & 62.7 & 51.3 & 52.2 & 59.3 & 54.9 \\

  Qwen3.5-9B
  & 55.7 & 49.7 & 59.7 & 61.6 & 44.4 & 60.0 & 62.4 & 55.6 \\

  Qwen3.5-397B-A13B
  & 61.9 & 58.8 & 56.3 & 65.3 & 68.8 & 62.0 & 65.4 & 62.0 \\

    GPT-5.4
    & 58.3 & 53.0 & 56.1 & 64.5 & 71.8 & 62.0 & 64.2 & 60.1 \\

    Qwen3.7-Plus
    & 60.5 & 58.8 & 60.5 & 65.9 & 72.6 & 64.9 & 61.1 & 62.6 \\

    Gemini-3.1-Pro
    & 63.6 & 62.1 & 58.1 & 67.4 & 74.4 & 67.0 & 67.9 & 64.8 \\

    \midrule

    \rowcolor{gray!15}
    \multicolumn{9}{l}{\textbf{\sys (Ours)}} \\
    \midrule

    \quad w/ Qwen3.5-397B-A13B
    & \textbf{75.5} & 65.8 & 65.2 & 76.1 & \textbf{79.5} & 58.3 & 80.2 & 72.1 \\

    \quad w/ Gemini-2.5-Pro
    & 69.4 & \textbf{76.5} & \underline{74.8} & 75.4 & \underline{78.3} & 61.6 & 73.8 & 72.8 \\

    \quad w/ Gemini-3.1-Pro
    & 73.8 & \underline{75.8} & \textbf{77.6} & \underline{79.0} & 76.1 & \textbf{68.8} & \underline{79.6} & \textbf{75.6} \\

    \quad w/ GPT-5.4
    & \underline{72.7} & 73.2 & 68.8 & \textbf{79.9} & 74.4 & \underline{67.0} & \textbf{82.7} & \underline{73.5} \\

    \midrule
    
    \quad \sys-4B
    & 61.0\textsubscript{\textcolor{green!50!black}{(+6.4)}}
    & 57.3\textsubscript{\textcolor{green!50!black}{(+5.3)}}
    & 53.5\textsubscript{\textcolor{red!75!black}{(-1.1)}}
    & 63.8\textsubscript{\textcolor{green!50!black}{(+1.1)}}
    & 58.1\textsubscript{\textcolor{green!50!black}{(+6.8)}}
    & 53.0\textsubscript{\textcolor{green!50!black}{(+0.8)}}
    & 69.1\textsubscript{\textcolor{green!50!black}{(+9.8)}}
    & 59.1\textsubscript{\textcolor{green!50!black}{(+4.2)}} \\
    
    \bottomrule
    \end{tabular}
    }
\end{table*}

\subsection{Performance across Document Types.}
Table~\ref{tab:domain_results} evaluates \sys across seven document types on MMLongBench-Doc.
\sys achieves the best results in every category, improving overall accuracy by 10.1--13.4 percentage points over the corresponding backbones.
Substantial gains span multiple document types: GPT-5.4 improves by 20.2 points on academic papers and 18.5 points on industry documents, while Gemini-3.1-Pro gains 19.5 points on guidebooks.
These improvements demonstrate that the benefits of interactive document solving extend beyond a single domain.

The strongest backbone nevertheless varies by document type.
Gemini-3.1-Pro achieves the highest overall accuracy (75.6), while GPT-5.4 leads on tutorials and industry documents, and Qwen3.5-397B leads on research and finance.
Gains are also less consistent on brochures: Gemini-3.1-Pro and GPT-5.4 improve by 1.8 and 5.0 points, respectively, whereas Qwen3.5-397B declines from 62.0 to 58.3.
This variation highlights the broad applicability of \sys while identifying document-specific limitations that remain sensitive to the underlying backbone.

\sys-4B improves overall accuracy from 54.9 to 59.1, with gains across six of seven document types.
The largest improvements occur on industry (+9.8), finance (+6.8), and research (+6.4), extending the benefits of interactive document solving to compact models.

\section{Case Study}
\label{sec:case_study}
% 不同类型的问题各来一个，然后带上不同baseline的错误结果

We examine three representative cases of \sys-4B after curriculum-based training, covering numerical computation, fine-grained visual understanding, and cross-page semantic comparison.
These trajectories illustrate how the trained agent composes individual capabilities according to the question and intermediate observations.

\begin{figure}[h!]
    \centering
    \includegraphics[width=\linewidth]{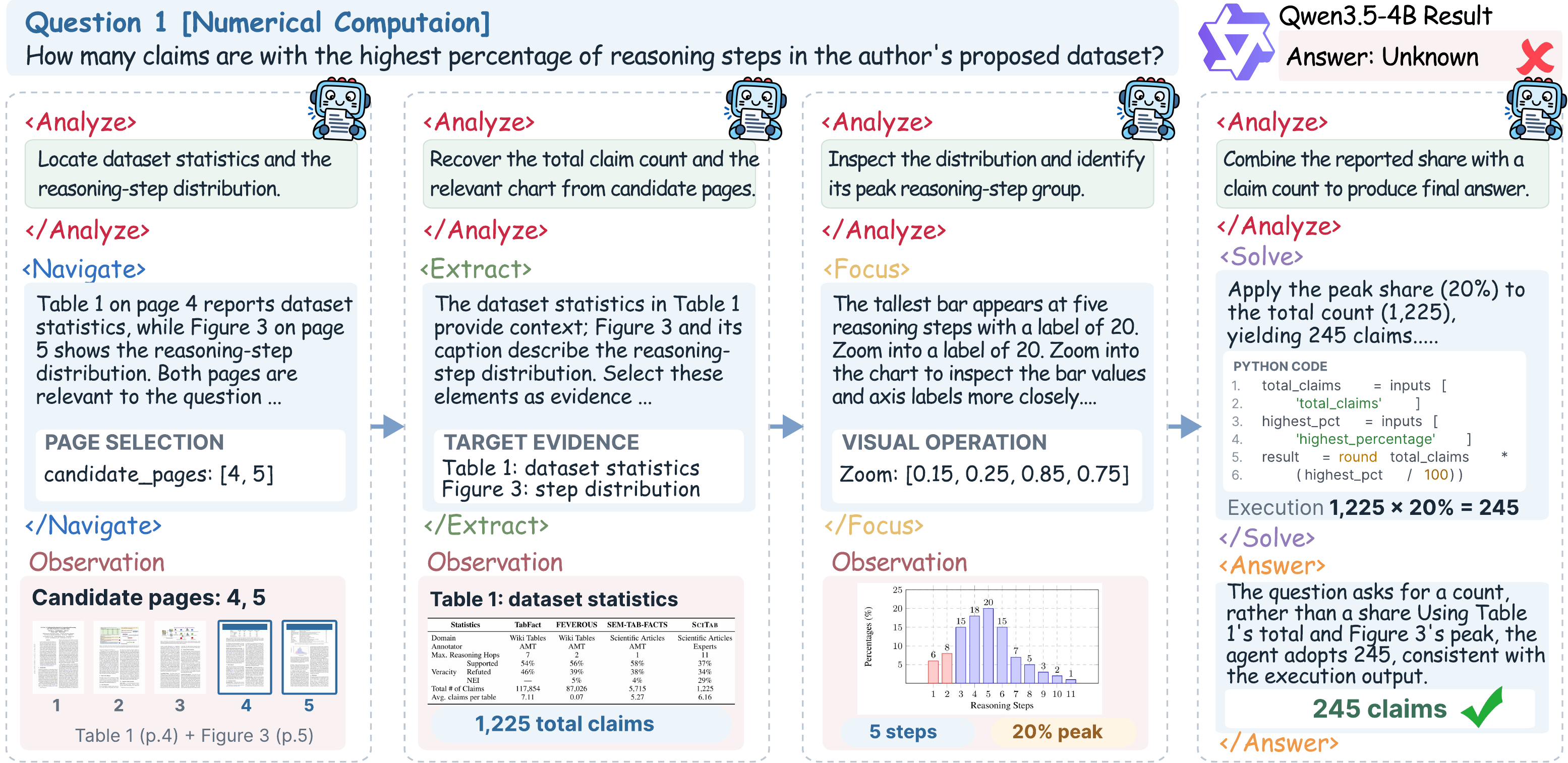}
    \vspace{-1em}
    \caption{A numerical computation case study showing \sys-4B's orchestration.}
    \label{fig:case_computation}
    \vspace{-1em}
 \end{figure}

\noindent\textbf{Numerical Computation.}
In Fig.~\ref{fig:case_computation}, the question requires combining dataset statistics with a distribution chart to determine a claim count.
\sys-4B uses \texttt{<Navigate>} and \texttt{<Extract>} to locate the relevant table and figure, then invokes \texttt{<Focus>} to inspect the chart before proceeding to \texttt{<Solve>}.
The resulting numerical reasoning yields a response of 245, whereas vanilla Qwen3.5-4B returns \texttt{unknown}.
This case illustrates the composition of evidence seeking, visual interpretation, and numerical reasoning: the agent first resolves the required quantities, then uses them to formulate the calculation.

\begin{figure}[p]
    \centering

    \includegraphics[
        width=\linewidth,
        height=0.28\textheight,
        keepaspectratio
    ]{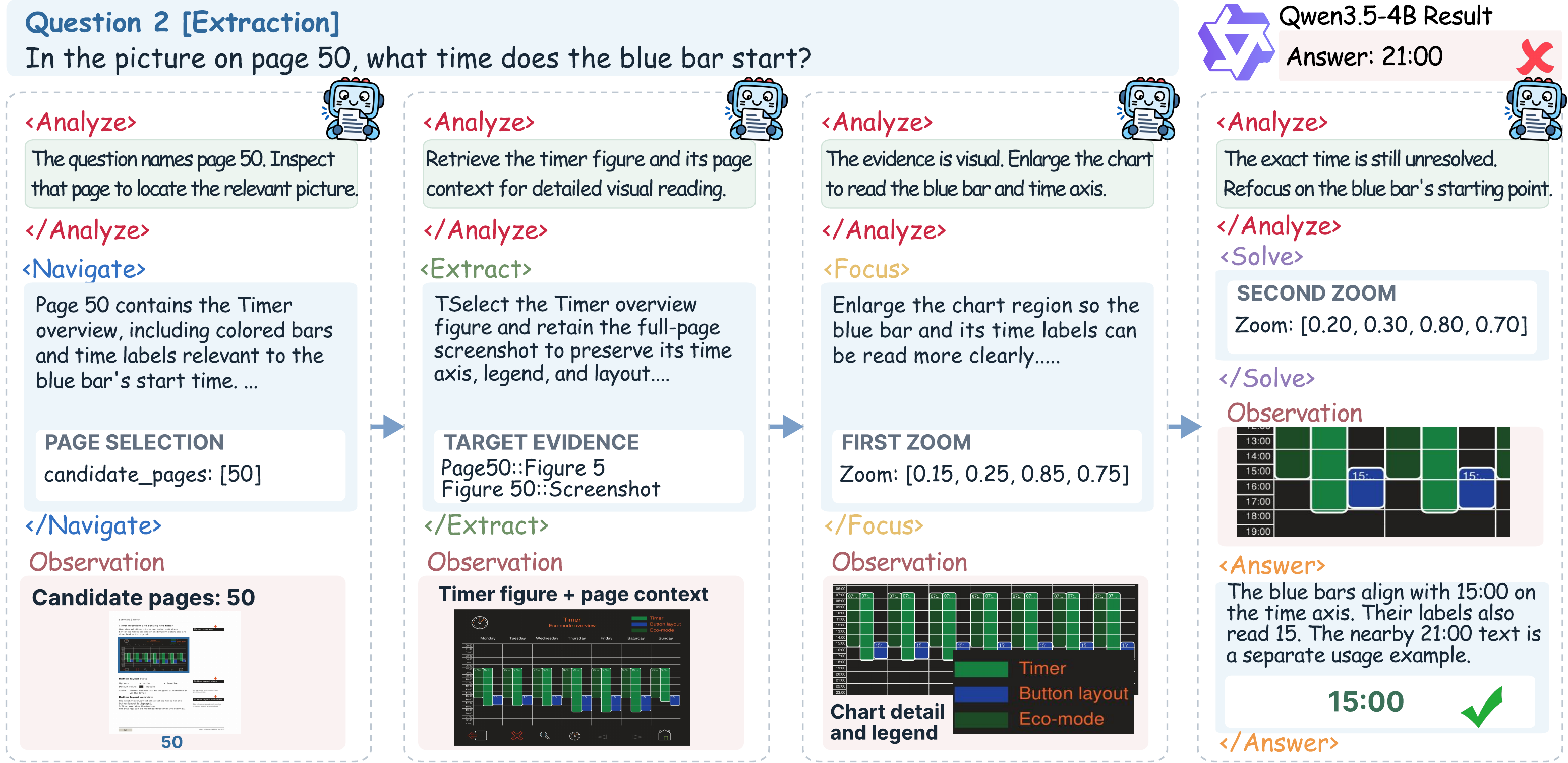}
    \vspace{-0.5em}
    \caption{An evidence extraction case study showing
    \sys-4B's orchestration.}
    \label{fig:case_extract}

    \vspace{0.5em}

    \includegraphics[
        width=\linewidth,
        height=0.3\textheight,
        keepaspectratio
    ]{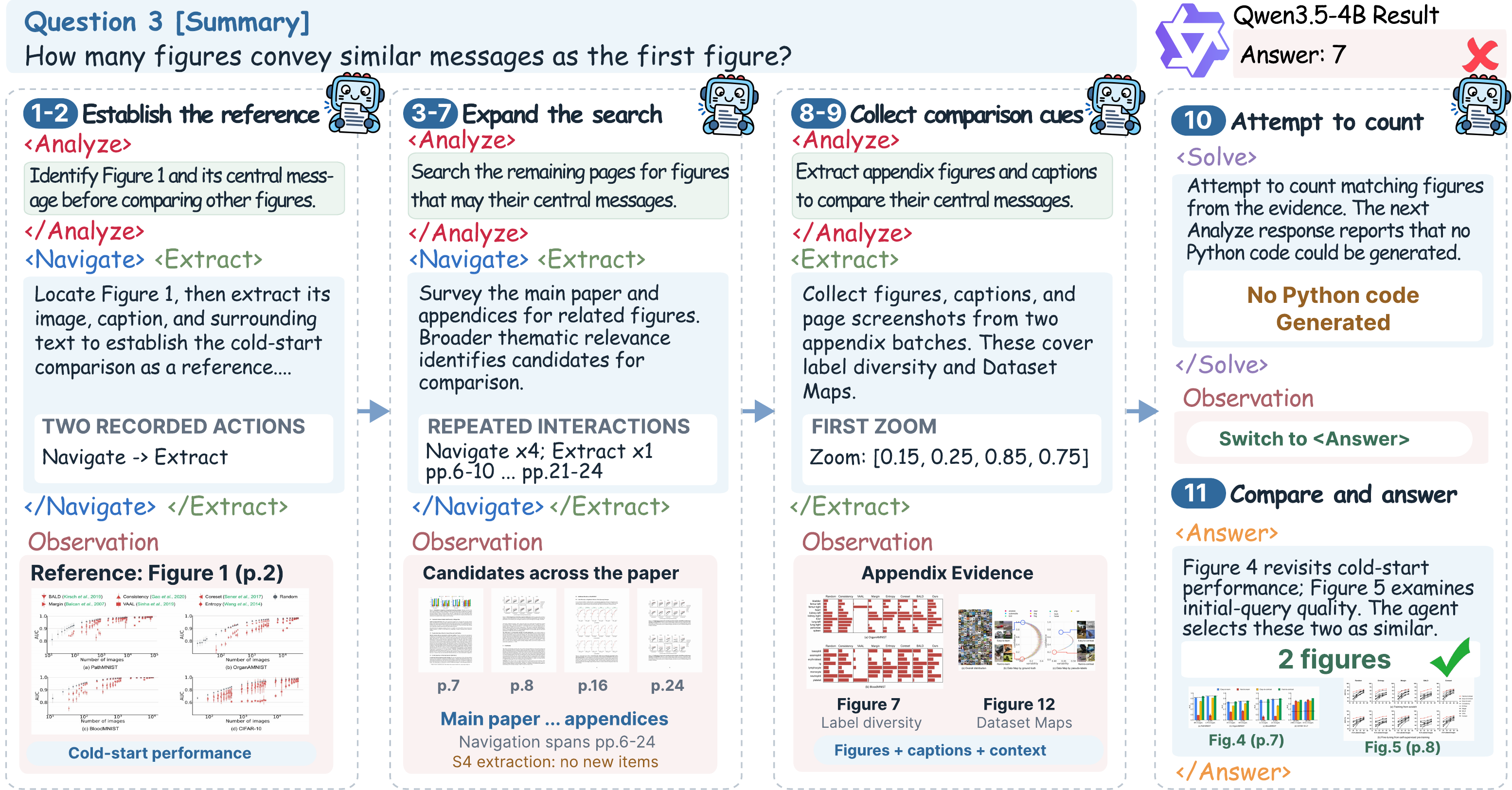}
    \vspace{-0.5em}
    \caption{A counting summary case study showing
    \sys-4B's orchestration.}
    \label{fig:case_summary}

    \vspace{0.5em}

    \includegraphics[
        width=\linewidth,
        height=0.32\textheight,
        keepaspectratio
    ]{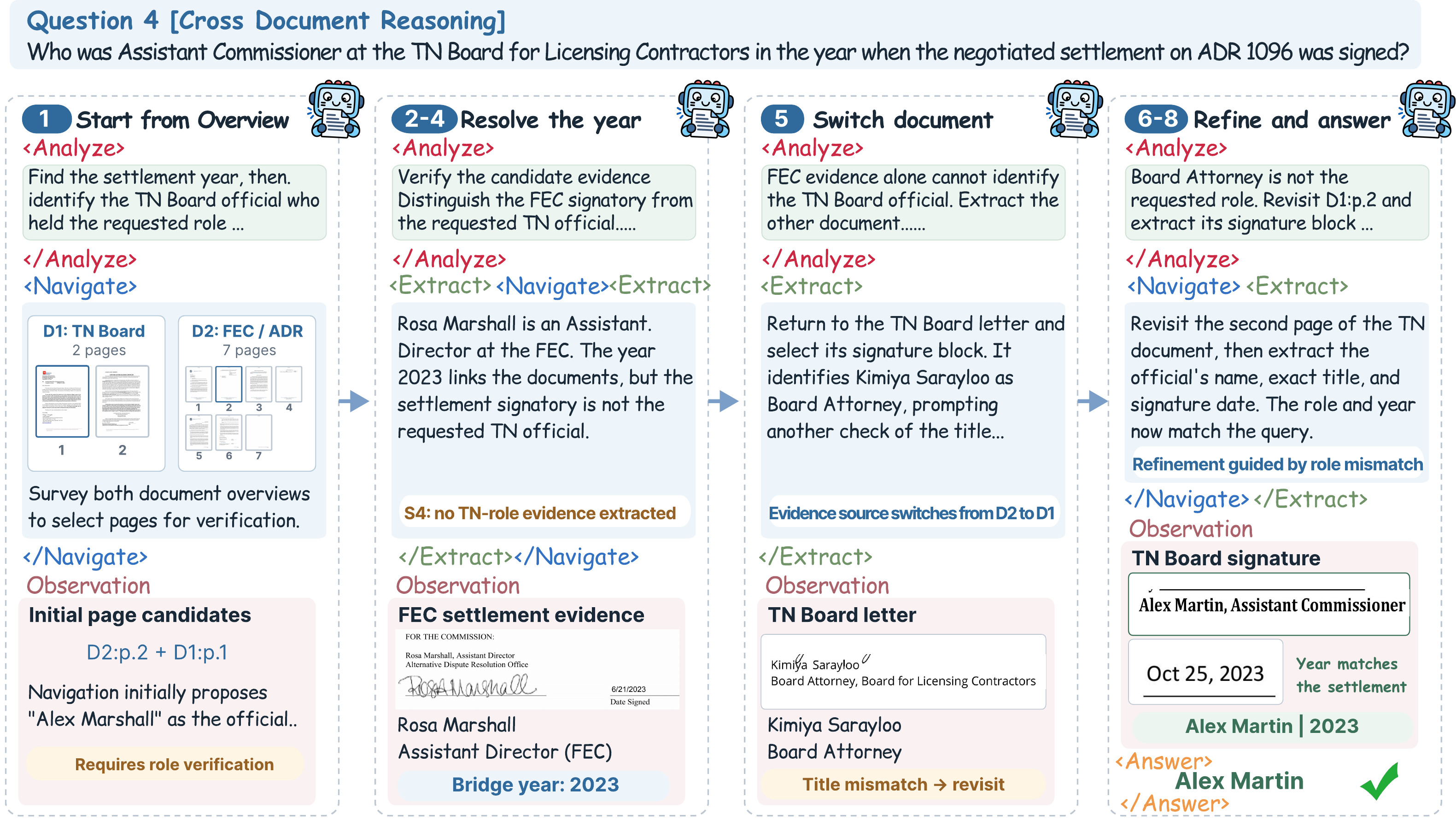}
    \vspace{-0.5em}
    \caption{A cross-document reasoning case study showing
    \sys-4B's orchestration.}
    \label{fig:case_crossdoc}
\end{figure}

\noindent\textbf{Fine-Grained Visual Understanding.}
Fig.~\ref{fig:case_extract} asks when a blue bar begins in a timer chart.
After locating the specified page and extracting the figure, \sys-4B performs two successive \texttt{<Focus>} operations to inspect the bar and its time label.
It correctly answers 15:00, distinguishing the visual evidence from a nearby textual example mentioning 21:00, which matches the vanilla model's incorrect response.
The case highlights the importance of adjusting inspection granularity: page-level relevance alone does not resolve a question whose answer depends on a small visual detail.

\noindent\textbf{Cross-Page Semantic Comparison.}
In Fig.~\ref{fig:case_summary}, the agent must count figures conveying a message similar to the first figure.
\sys-4B establishes the reference message, then repeatedly invokes \texttt{<Navigate>} and \texttt{<Extract>} to collect comparison evidence across the main paper and appendices.
Following an unsuccessful \texttt{<Solve>} attempt, \texttt{<Analyze>} redirects the process to \texttt{<Answer>} for semantic comparison, ultimately selecting Figures 4 and 5.
This trajectory illustrates sustained evidence acquisition and adjustment after tool failure, extending orchestration beyond a fixed sequence of successful actions.

\noindent\textbf{Cross-Document Reasoning.}
In Fig.~\ref{fig:case_crossdoc}, \sys-4B starts from document overviews and invokes \texttt{<Navigate>} and \texttt{<Extract>} to gather evidence from the FEC settlement and TN Board documents.
\texttt{<Analyze>} identifies mismatches between the retrieved titles and the requested role, prompting re-navigation to the TN Board's signature page.
Linking the settlement year, 2023, to the official's title and signature date yields Alex Martin.
This trajectory illustrates how intermediate observations guide document switching and role disambiguation to support cross-document evidence integration.

These distinct interaction patterns qualitatively support the curriculum's goal of connecting individual document-solving capabilities with adaptive orchestration.

 % \begin{figure}[h!]
 %    \centering
 %    \includegraphics[width=\linewidth]{fig/case_extract.pdf}
 %    \vspace{-1em}
 %    \caption{A evidence extraction case study showing \sys-4B's orchestration.}
 %    \label{fig:case_extract}
 %    \vspace{-2em}
 % \end{figure}

 % \begin{figure}[h!]
 %    \centering
 %    \includegraphics[width=\linewidth]{fig/case_summary.pdf}
 %    \vspace{-1em}
 %    \caption{A counting summary case study showing \sys-4B's orchestration.}
 %    \label{fig:case_summary}
 %    \vspace{-2em}
 % \end{figure}

 %  \begin{figure}[h!]
 %    \centering
 %    \includegraphics[width=\linewidth]{fig/case_crossdoc.pdf}
 %    \vspace{-1em}
 %    \caption{A cross-doc reasoning case study showing \sys-4B's orchestration.}
 %    \label{fig:case_crossdoc}
 % \end{figure}

 \clearpage
\clearpage

\FloatBarrier
\section{System Prompt Design}
\label{sec:app_tool_prompt}

\definecolor{cBlue_6}{HTML}{2B4B80}
\definecolor{cBlue_1}{HTML}{007AFF}
\definecolor{titleBlue1}{HTML}{3D7197}
\definecolor{contentBlue1}{HTML}{2A5574}
\definecolor{titleBlue2}{HTML}{8599AD}
\definecolor{contentBlue2}{HTML}{2A5574}

\begin{tcolorbox}[
    width=\linewidth,
    nobeforeafter,
    colback=cBlue_1!2,
    colframe=titleBlue2,
    colbacktitle=titleBlue2,
    coltitle=white,
    title=\textbf{Navigate Agent Prompt},
    fonttitle=\bfseries\small,
    fontupper=\footnotesize,
    boxrule=0.4mm,
    arc=1.5mm,
    left=4pt,
    right=4pt,
    top=4pt,
    bottom=4pt,
    enhanced,
    breakable
]

\noindent\textbf{[System Role \& Objective]}\\
You are the Navigate Agent in a Document Visual Question Answering (DocVQA) system.
Your objective is to scan document pages and identify \textbf{ANY} page with potential to contribute to answering the search query.

\vspace{0.5em}

\noindent\textbf{[Crucial Constraint]}\\
You are \textbf{EXCLUSIVELY} a page-level navigator.\\
Do \textbf{NOT} attempt to answer the user's question.\\
Do \textbf{NOT} suggest downstream tools, extraction strategies, or region-level focus operations.

\vspace{0.5em}

% \noindent\textbf{[Modality Compliance]}\\
% Strictly respect the configured \texttt{modal} setting.\\
% \texttt{text}: Evaluate text-only context.\\
% \texttt{image}: Evaluate page screenshots or thumbnails.\\
% \texttt{text,image} (or \texttt{[text, image]}): Synthesize both textual content and visual layouts.

\vspace{0.5em}

\noindent\textbf{[Page Selection \& Navigation Strategy]}\\
\textbf{1. High Recall for NEW Pages:}
Include any new page containing exact keywords, synonyms, related thematic concepts, or structural elements such as tables or figures that might hold the answer. Do not discard new pages using a rigid confidence threshold; use the \texttt{confidence} score to rank relevance.\\
\textbf{2. High Precision for PREVIOUS Candidates:}
Re-evaluate previously identified candidate pages rigorously and retain only those with strong ongoing relevance.\\
\textbf{3. Avoid Redundant Work:}
Do not re-select pages that have already been fully verified by current evidence.\\
\textbf{4. Multi-Hop Look-Ahead:}
Set \texttt{need\_more\_pages=true} when current evidence is incomplete or unvisited pages may contain missing evidence for multi-step reasoning.\\
\textbf{5. Empty Selection:}
If no current page is relevant and no previous candidate remains strongly relevant, return an empty \texttt{candidate\_pages} array and explain why in \texttt{reason}.\\
\textbf{6. No Out-of-Range Pages:}
Only select page numbers appearing in the input \texttt{page\_set} or \texttt{pages}. Never invent or extrapolate page numbers.\\
\textbf{7. Unavailable Target Pages:}
If the query requests a specific page that is absent from the current input, return \texttt{candidate\_pages: []}, explain that the page is unavailable, and do not substitute nearby pages unless explicitly relevant to the same evidence need.

\vspace{0.5em}

\noindent\textbf{[Strict Output Constraints]}\\
\textbf{Page Format:} Use integer page numbers only.\\
\textbf{Page Boundary:} Every selected page must appear in the input \texttt{page\_set}.\\
\textbf{Output Schema:} Return only a valid JSON object matching the format below,

\vspace{0.4em}

\begin{verbatim}
{
  "candidate_pages": [
    {
      "page": 1,
      "confidence": 0.85
    }
  ],
  "reason": "brief explanation of why the selected pages are relevant",
  "need_more_pages": true
}
\end{verbatim}

\end{tcolorbox}

\begin{tcolorbox}[
    width=\linewidth,
    nobeforeafter,
    colback=cBlue_1!2,
    colframe=titleBlue2,
    colbacktitle=titleBlue2,
    coltitle=white,
    title=\textbf{Extract Agent Prompt},
    fonttitle=\bfseries\small,
    fontupper=\footnotesize,
    boxrule=0.4mm,
    arc=1.5mm,
    left=4pt,
    right=4pt,
    top=4pt,
    bottom=4pt,
    enhanced,
    breakable
]

\noindent\textbf{[System Role \& Objective]}\\
You are the Extract Agent in a Document Visual Question Answering (DocVQA) system. \\
Your objective is to evaluate candidate items from selected pages and retain only those containing specific, actionable evidence required to answer the search query.

\vspace{0.5em}

\noindent\textbf{[Input Context]}\\
You will receive the \textbf{search query} and a \textbf{candidate list} containing natural-language descriptions of Text, Table, Figure, and page screenshot nodes.

\vspace{0.5em}

\noindent\textbf{[Core Selection Strategy]}\\
\textbf{1. Answerability is Key:}
Select only items containing the facts, values, qualifiers, labels, definitions, or visual/structural evidence required to resolve the query. Reject generic or tangential content.\\
\textbf{2. Evidence Completeness:}
Ensure the selected evidence covers all required answer slots, including entity or category, metric, time period, units, labels, and calculation inputs. If necessary, select the smallest combination of items that provides complete support.\\
\textbf{3. Precision with Coverage:}
Keep the selection focused while retaining evidence required to interpret the answer. If local nodes may omit important layout, legends, paired figures, or cross-node relationships, also select the full-page screenshot as supporting visual evidence.\\
\textbf{4. Mixed Evidence:}
Select evidence across modalities when the query requires cross-referencing text, tables, figures, captions, or legends. Do not assume numerical questions can be resolved from tables alone when surrounding text provides necessary context or constraints.\\
\textbf{5. Empty Return:}
If none of the candidates provide useful evidence, return an empty list.

\vspace{0.5em}

\noindent\textbf{[Modality-Specific Prioritization]}\\
\textbf{Visual \& Spatial Queries:}
Prioritize relevant Figure nodes or page screenshots. For multi-panel figures, select the containing Figure node for downstream subfigure inspection.\\
\textbf{Textual Queries:}
Prioritize relevant Text nodes with definitions, context, quotes, or facts.\\
\textbf{Tabular Queries:}
Prioritize relevant Table nodes for lookups, comparisons, totals, or rankings.\\
\textbf{Mixed Queries:}
Select evidence across modalities when jointly required.

\vspace{0.5em}

\noindent\textbf{[Strict Output Constraints]}\\
Return only a valid JSON object matching the schema below.\\
Do \textbf{NOT} include Markdown formatting or any text outside the JSON object.

\vspace{0.4em}

\begin{verbatim}
{
  "selected_candidate_ids": [
    "page7::text::7",
    "page7::screenshot"
  ],
  "reason": "A brief, one-sentence explanation of how the selected items support the query."
}
\end{verbatim}

\end{tcolorbox}

\begin{tcolorbox}[
  width=\linewidth,
  nobeforeafter,
  colback=cBlue_1!2,
  colframe=titleBlue2,
  colbacktitle=titleBlue2,
  coltitle=white,
  title=\textbf{Focus Agent Prompt},
  fonttitle=\bfseries\small,
  fontupper=\footnotesize,
  boxrule=0.4mm,
  arc=1.5mm,
  left=4pt,
  right=4pt,
  top=4pt,
  bottom=4pt,
  enhanced,
  breakable
]

\noindent\textbf{[System Role \& Objective]}\\
You are the Focus Planner in a Document Visual Question Answering (DocVQA) system.\\
Your objective is to plan high-precision visual reading operations for an already extracted figure.

\vspace{0.5em}

\noindent\textbf{[Crucial Constraint]}\\
Output only visual operations. Do \textbf{NOT} attempt to answer the user's question directly.

\vspace{0.5em}

\noindent\textbf{[Allowed Operations]}
\begin{itemize}[leftmargin=*, labelsep=0.5em, topsep=0pt, itemsep=2pt]
  \item \texttt{subfigure\_split}: Use for multi-panel figures. Specify the target panel in \texttt{region.panel\_query} and provide \texttt{grid\_hint} when the layout is clear. Do not combine it with full-figure crop or zoom.

  \item \texttt{crop}: Use for clearly bounded regions such as source notes, footer text, or distinct textual blocks.

  \item \texttt{zoom}: Use for small labels, legends, axis labels, or exact numerical values.
\end{itemize}

\vspace{0.5em}

\noindent\textbf{[Region Specification]}\\
Specify the target region in the \texttt{region} object. Prefer normalized coordinates \texttt{[x1, y1, x2, y2]} in the $[0,1]$ figure-relative space when the region is geometrically clear; otherwise, use a descriptive coarse region.

\vspace{0.5em}

\noindent\textbf{[Visual Captioning]}\\
The \texttt{visual\_caption} is required and should briefly summarize the localized visual content and relevant entities, labels, values, trends, panel identities, or legend mappings.\\
Do \textbf{NOT} state the final answer unless it is directly visible in the selected region, and avoid instructional descriptions such as ``inspect labels'' or ``zoom into the chart.''

\vspace{0.5em}

\noindent\textbf{[Output Format]}\\
Return only valid JSON following the schema of the selected operation. Use the minimum number of visual operations required for the current focus goal, and do not combine incompatible operations.

\end{tcolorbox}

\begin{tcolorbox}[
  width=\linewidth,
  nobeforeafter,
  colback=cBlue_1!2,
  colframe=titleBlue2,
  colbacktitle=titleBlue2,
  coltitle=white,
  title=\textbf{Solve Agent Prompt},
  fonttitle=\bfseries\small,
  fontupper=\footnotesize,
  boxrule=0.4mm,
  arc=1.5mm,
  left=4pt,
  right=4pt,
  top=4pt,
  bottom=4pt,
  enhanced,
  breakable
]

\noindent\textbf{[System Role \& Objective]}\\
You are the Solver Agent in a Document Visual Question Answering (DocVQA) system.\\
Your objective is to generate robust Python code for deterministic computation over extracted structured evidence, especially tabular data.

\vspace{0.5em}

\noindent\textbf{[Execution Mechanism]}\\
The generated \texttt{code} is executed by the environment using variables provided in \texttt{inputs}.\\
The final computed answer must be assigned to the Python variable \texttt{result}.

\vspace{0.5em}

\noindent\textbf{[Coding Rules]}\\
\textbf{1. Variable Requirement:}
Always assign the final output to \texttt{result}.\\
\textbf{2. Input Dependency:}
Read data only from the \texttt{inputs} dictionary.\\
\textbf{3. Allowed Imports:}
Only \texttt{pandas}, \texttt{numpy}, \texttt{re}, and \texttt{math} may be imported. Do not access files, APIs, or the network.\\
\textbf{4. Determinism:}
Keep all computation deterministic and reproducible.

\vspace{0.5em}

\noindent\textbf{[Data Processing Standards]}\\
\textbf{Data Cleaning:}
Normalize formatted values such as currencies, percentages, missing values, and accounting-style negatives before computation.\\
\textbf{DataFrame Processing:}
For tabular evidence, store records in \texttt{inputs} and convert them to a Pandas DataFrame for filtering, aggregation, or comparison.\\
\textbf{Error Handling:}
Guard against empty results, invalid values, and division by zero.\\
\textbf{Reported Quantity:}
Use the explicitly reported values requested by the question. Do not reconstruct alternative quantities unless the target value is absent and the evidence explicitly defines the derivation.

\vspace{0.5em}

\noindent\textbf{[Fallback]}\\
If the evidence is insufficient for grounded computation, return:

\begin{verbatim}
{
"summary": "Evidence is insufficient, so the solver returns result = None.",
"code": "result = None",
"inputs": {}
}
\end{verbatim}

\vspace{0.5em}

\noindent\textbf{[Strict Output Constraints]}\\
Return only a valid JSON object containing \texttt{summary}, \texttt{code}, and \texttt{inputs}.\\
Do not include Markdown formatting or additional text. Python newlines in \texttt{code} must be escaped as \texttt{\textbackslash n}.

\end{tcolorbox}

\begin{tcolorbox}[
  width=\linewidth,
  nobeforeafter,
  colback=cBlue_1!2,
  colframe=titleBlue2,
  colbacktitle=titleBlue2,
  coltitle=white,
  title=\textbf{Answer Agent Prompt},
  fonttitle=\bfseries\small,
  fontupper=\footnotesize,
  boxrule=0.4mm,
  arc=1.5mm,
  left=4pt,
  right=4pt,
  top=4pt,
  bottom=4pt,
  enhanced,
  breakable
]

\noindent\textbf{[System Role \& Objective]}\\
You are the unified Answer Agent in a Document Visual Question Answering (DocVQA) pipeline.\\
Your objective is to generate a concise, factual answer based \textbf{EXCLUSIVELY} on the provided evidence.

\vspace{0.5em}

\noindent\textbf{[Grounding \& Anti-Hallucination]}\\
\textbf{Strict Grounding:}
Rely entirely on the provided evidence. Do not invent values, entities, relationships, or assumptions.\\
\textbf{Page Grounding:}
Mention only page numbers explicitly present in the evidence.

\vspace{0.5em}

\noindent\textbf{[Reasoning \& Deduction]}\\
\textbf{Rigorous Reasoning:}
Apply the question criteria systematically and connect all required evidence into a complete reasoning chain.\\
\textbf{Semantic Alignment:}
Match the target entity, metric, time period, unit, and comparison scope to the evidence. Return ``Not Answerable'' when explicit constraints conflict.\\
\textbf{Use the Reported Quantity:}
Use values reported under the labels named in the question. Reconstruct quantities only when the requested value is absent and the document explicitly defines the derivation.\\
\textbf{Count vs. Share:}
Return percentages only when explicitly requested. For headcount questions, compute from the relevant sample size and rate; return ``INSUFFICIENT\_EVIDENCE'' if the required sample size is unavailable.\\
\textbf{Consistency:}
The final answer must follow directly from the reasoning in the rationale.

\vspace{0.5em}

\noindent\textbf{[Fallback Rules]}\\
\textbf{Deterministic Derivation:}
When all required numerical inputs are available, compute the requested value directly rather than abstaining.\\
\textbf{Document-provided Fallback:}
Use an explicit default, remaining, or fallback value when the requested item is not separately listed but the document provides an applicable alternative.\\
\textbf{Incomplete Evidence:}
Return ``INSUFFICIENT\_EVIDENCE'' only when a required fact is missing and no deterministic derivation or document-provided fallback is available.\\
\textbf{Exactness Limit:}
If an exact answer is requested but the evidence supports only a vague visual estimate, return ``Not Answerable''.

\vspace{0.5em}

\noindent\textbf{[Output Format]}\\
Return only a valid JSON object matching the schema below,

\begin{verbatim}
{
"answer": "The concise final answer, or a fallback string (Not Answerable).",
"rationale": "A brief evidence-based explanation.",
"confidence": 0.95
}
\end{verbatim}

\end{tcolorbox}

\begin{tcolorbox}[
  width=\linewidth,
  nobeforeafter,
  colback=cBlue_1!2,
  colframe=titleBlue2,
  colbacktitle=titleBlue2,
  coltitle=white,
  title=\textbf{Analyze Agent Prompt},
  fonttitle=\bfseries\small,
  fontupper=\footnotesize,
  boxrule=0.4mm,
  arc=1.5mm,
  left=4pt,
  right=4pt,
  top=4pt,
  bottom=4pt,
  enhanced,
  breakable
]

\noindent\textbf{[System Role \& Objective]}\\
You are the Textual Analysis Agent within the DocPilot system.\\
Your sole responsibility is to verify whether the current candidate answer is fully supported by the collected evidence and aligned with the original question. You reason only over the provided textual state and do not directly execute tools.

\vspace{0.5em}

\noindent\textbf{[General Rules]}\\
\textbf{Strict Grounding:}
Base verification exclusively on the provided question, evidence, and current state. Do not introduce external knowledge.\\
\textbf{Answer Type:}
Infer the expected answer type from the question, such as entity, percentage, ratio, or count, and verify that the candidate matches it.\\
\textbf{Page Grounding:}
Mention only page numbers explicitly present in the supporting evidence.\\
\textbf{Support vs. Coverage:}
Distinguish whether the candidate is supported by current evidence from whether the search coverage is sufficient. Acceptance requires both.

\vspace{0.5em}

\noindent\textbf{[Failure Diagnosis \& Tool Mapping]}\\
\texttt{navigate}: Use when relevant pages may remain unexplored or page-level coverage is incomplete.\\
\texttt{extract}: Use when the correct page is known but required text, table, figure, or related nodes are missing.\\
\texttt{focus}: Use when visual evidence is available but labels, subfigures, or spatial details remain unresolved.\\
\texttt{solve}: Use when evidence is available but deterministic computation, or comparison remains unresolved.\\
\texttt{answer}: Use when evidence is sufficient but the candidate answer is poorly synthesized or misaligned.

\vspace{0.5em}

\noindent\textbf{[Verification Criteria]}\\
Accept the candidate when it is grounded, internally consistent, and covers the requested scope.\\
Reject only for clear fatal flaws, such as inconsistent reasoning, unsupported conclusions, incomplete comparisons, incorrect counts, or missing required evidence.\\
Do not reject based on speculation, unused contextual facts, or generic uncertainty.

\vspace{0.5em}

\noindent\textbf{[Unanswerable Cases]}\\
\texttt{Not Answerable} may be accepted when the evidence sufficiently establishes that the requested information is unavailable, outside the document scope, or not applicable.

\vspace{0.5em}

\noindent\textbf{[Output Format]}\\
Return only a valid JSON object matching the schema below,

\begin{verbatim}
{
"result": true,
"issues": "Empty when accepted; otherwise state the decisive flaw or evidence gap.",
"suggestion": "Empty when accepted; otherwise list next action",
"summary": "Briefly summarize the verification result."
}
\end{verbatim}

\end{tcolorbox}

\end{document}